\documentclass[10pt,journal,compsoc]{IEEEtran}
\ifCLASSOPTIONcompsoc
\else
\fi
\ifCLASSINFOpdf
\else
\fi
\ifCLASSOPTIONcompsoc
\usepackage[normalsize,sf]{subfigure}
\else
\usepackage[footnotesize]{subfigure}
\fi

\ifCLASSOPTIONcompsoc
  \usepackage[caption=false,font=normalsize,labelfont=sf,textfont=sf]{subfig}
\else1
  \usepackage[caption=false,font=footnotesize]{subfig}
\fi
\usepackage{times}
\usepackage{epsfig}
\usepackage{graphicx}
\usepackage{amsmath}
\usepackage{amssymb}
\usepackage{subfigure}
\usepackage{multirow}
\usepackage[algoruled,vlined,linesnumbered]{algorithm2e}
\usepackage{rotating}
\usepackage{bbm}
\usepackage{bm}
\usepackage{color}
\usepackage[dvipsnames]{xcolor}
\usepackage{threeparttable}
\usepackage{multirow}
\usepackage{rotating}
\usepackage{makecell}
\usepackage{geometry}
\newcommand{\wsf}[1]{{\color{black}#1}}
\newcommand{\zhu}[1]{{\color{black}#1}}
\newcommand{\zhux}{\textcolor[rgb]{0,0,0}}
\newcommand{\zhuxt}{\textcolor[rgb]{0,0,0}}
\newcommand{\cyc}{\textcolor[rgb]{0,0,0}}
\newcommand{\cycb}{\textcolor[rgb]{0,0,0}}
\newcommand{\cycc}{\textcolor[rgb]{0,0,0}}
\newcommand{\cycd}{\textcolor[rgb]{0,0,0}}
\newcommand{\wsff}[1]{{\color{black}#1}}

\newcommand{\xpami}[1]{{\color{black}#1}}

\newcommand{\db}{\textcolor[rgb]{0,0,1}}
\newcommand{\dr}{\textcolor[rgb]{1.00,0.00,0.00}}

\usepackage[nospace,nocompress]{cite}

\usepackage{cancel}

\usepackage{algorithmic}

\begin{document}

\newgeometry{top=6cm,bottom=1cm}

\onecolumn{


\noindent \textbf{\Huge{Person Re-Identification by Camera \\ Correlation Aware Feature Augmentation}}

\vspace{2cm}

\noindent {\LARGE{Ying-Cong Chen, Xiatian Zhu, Wei-Shi Zheng, Jian-Huang Lai}}

\Large

\vspace{2cm}

\noindent Code is available at the project page: \\ 
\ \ \ \ \ \ \ \ \ \ \ \ {http://isee.sysu.edu.cn/\%7ezhwshi/project/CRAFT.html}

\vspace{1cm}

\noindent For reference of this work, please cite:

\vspace{1cm}

\noindent Ying-Cong Chen, Xiatian Zhu, Wei-Shi Zheng, and Jian-Huang Lai. Person Re-Identification by Camera Correlation Aware Feature Augmentation. IEEE  Transactions on Pattern Analysis and Machine Intelligence (DOI: 10.1109/TPAMI.2017.2666805)

\vspace{1cm}

\noindent Bib:

\noindent 
@article\{craft,\\
\ \ \   title=\{Person Re-Identification by Camera Correlation Aware Feature Augmentation\},\\
\ \ \  author=\{Chen, Ying-Cong and Zhu, Xiatian and Zheng, Wei-Shi and Lai, Jian-Huang\},\\
\ \ \  journal=\{IEEE  Transactions on Pattern Analysis and Machine Intelligence (DOI: 10.1109/TPAMI.2017.2666805)\}\\
\}

}

  
 \clearpage  
  
\restoregeometry

%
\title{Person Re-Identification by Camera Correlation Aware Feature Augmentation}
%
%
%
%

\author{
Ying-Cong Chen,
Xiatian Zhu,
Wei-Shi Zheng,
Jian-Huang Lai
\IEEEcompsocitemizethanks{
\IEEEcompsocthanksitem
Ying-Cong Chen is with the School of Data and Computer Science, Sun
Yat-sen University, Guangzhou 510275, China, with the Collaborative
Innovation Center of High Performance Computing, National University
of Defense Technology, Changsha 410073, China, and is also with Department
of Computer Science and Engineering, The Chinese University
of Hong Kong. E-mail: yingcong.ian.chen@gmail.com
\protect
\IEEEcompsocthanksitem 
Xiatian Zhu is with School of Data and Computer Science, Sun Yat-sen
University, Guangzhou 510275, China; and also with School of Electronic
Engineering and Computer Science, Queen Mary University of London,
United Kingdom. E-mail: xiatian.zhu@qmul.ac.uk
\protect
\IEEEcompsocthanksitem 
Wei-Shi Zheng is with the School of Data and Computer Science, Sun
Yat-sen University, Guangzhou 510275, China., and is also with the Key
Laboratory of Machine Intelligence and Advanced Computing (Sun Yatsen
University), Ministry of Education, China. E-mail: wszheng@ieee.org.
\protect
\IEEEcompsocthanksitem 
Jian-Huang Lai are with the School of Data and Computer Science,
Sun Yat-sen University, Guangzhou 510275, China, and is also with
Guangdong Province Key Laboratory of Information Security, P. R. China.
E-mail: stsljh@mail.sysu.edu.cn.
}
}

\IEEEcompsoctitleabstractindextext{%
\begin{abstract}

The challenge of person re-identification (re-id) is to match individual images of the same person captured by different non-overlapping camera views against significant and unknown cross-view feature distortion. 
While a large number of distance metric/subspace learning models have been developed for re-id, the cross-view transformations they learned are view-generic and thus potentially less effective in quantifying the feature distortion inherent to each camera view. Learning view-specific feature transformations for re-id (i.e., view-specific re-id), an under-studied approach, becomes
an alternative
resort for this problem.
In this work,
we formulate a novel view-specific person re-identification framework 
from the feature augmentation point of view,
called
\textbf{C}amera co\textbf{R}relation \textbf{A}ware \textbf{F}eature augmen\textbf{T}ation
(CRAFT).
Specifically, CRAFT performs cross-view adaptation by automatically measuring camera correlation
from cross-view visual data distribution
and adaptively conducting feature augmentation to transform the original features
into a new adaptive space.
Through {our} augmentation framework, view-generic learning algorithms can be readily generalized to
learn and optimize view-specific sub-models whilst simultaneously modelling view-generic discrimination information.
Therefore, our framework not only inherits the strength of view-generic model learning but also provides an effective way to take into account view specific characteristics.
Our CRAFT framework can be extended to jointly learn view-specific feature transformations for person re-id across
a large network with more than two cameras, a largely under-investigated but realistic re-id setting. 
Additionally, we present a domain-generic deep person appearance representation
which is designed particularly to be towards view invariant for facilitating cross-view adaptation by CRAFT.
We conducted extensively comparative experiments to validate
the superiority and advantages of our proposed framework over state-of-the-art competitors
on contemporary challenging person re-id datasets.

\end{abstract}

}

\maketitle

\IEEEdisplaynotcompsoctitleabstractindextext

%

\IEEEpeerreviewmaketitle


\section{Introduction}
\label{sec:intro}

{\color{black}The
extensive deployment of close-circuit television cameras in visual surveillance
results in \zhux{a} vast quantity of visual data and
necessitates inevitably automated data interpretation mechanisms.
One of the most essential visual data processing tasks is to
automatically re-identify individual {person}
across non-overlapping camera views distributed at different physical locations,
which is known as {person re-identification} (re-id).
However, person re-id by visual matching is inherently challenging
due to {the} {existence of} many visually similar people and
dramatic appearance changes of the same {person}
arising from the great cross-camera variation in viewing conditions such as illumination, viewpoint, occlusions and background clutter \cite{gong2014person}
(Figure \ref{fig:challenge}).
%
%
%
%

In current person re-id literature,
the best performers are discriminative learning based methods
\cite{medFilter,colorName,colorInvariants,LMNN,PCCA,ITML,PRDC,LADF,LFDA,kernelREIDECCV14,XQDA,KISSME,ensembleReID,LiaoPSD,ROCCA,an2015person,lisanti2014matching,chen2015CVDCA,zheng2016towards,wang2015cross}.
Their essential objective is to establish a reliable re-id matching model
through learning identity discriminative information
from the pairwise training data.
Usually, this is achieved by either
{\em view-generic} modelling
(e.g., optimizing a common model for multiple camera views)
\cite{LMNN,PCCA,PRDC,LFDA,KISSME,LiaoPSD}
or 
{\em view-specific} modelling scheme
(e.g., optimizing a separate model for each camera view)
\cite{ROCCA,an2015person,lisanti2014matching,chen2015CVDCA}.
{The former mainly focuses on the shared view-generic discriminative learning but does not 
  explicitly take the individual view information ({e.g., via camera view labels}) into modelling.
Given that person re-id inherently incurs
dramatic appearance change across
camera views due to the great
difference in illumination, viewpoint or camera {characteristics}, 
the view-generic approach is inclined to be sub-optimal in 
quantifying the feature distortion caused by these {variations} in individual camera views. 
While the latter approach may enable to mitigate this problem by 
particularly considering view label information during modelling, 
most of these methods do not {explicitly} take into consideration 
the feature distribution alignment across camera views
so that cross-view data adaptation {cannot} be directly quantified and optimized during model learning
as view-generic counterparts do. 
Additionally, existing view-specific methods are often subject to limited scalability
for person re-id across multiple (more than two) camera views 
in terms of implicit assumptions and formulation design.}

\begin{figure}
	\centering
	\includegraphics[width=1\linewidth]{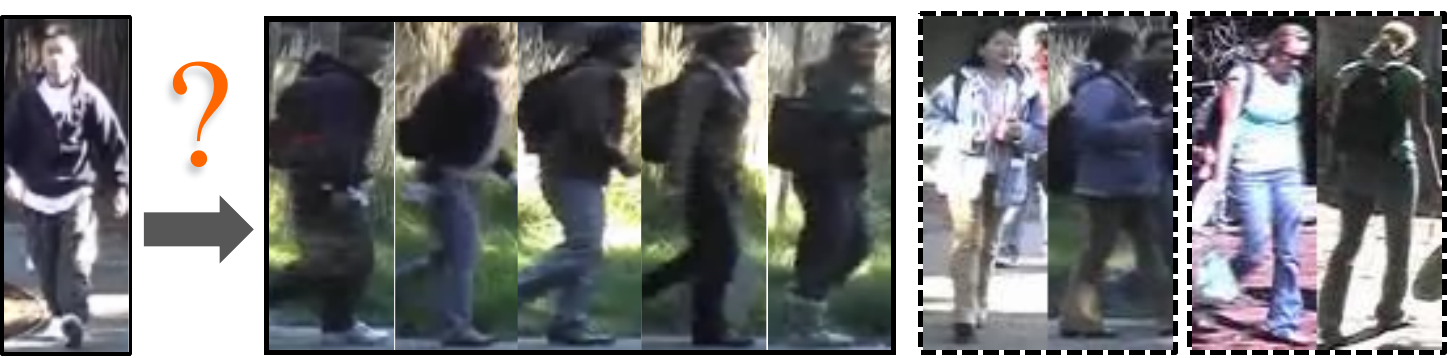}
	\vskip -.4cm
	\caption{\footnotesize
		Illustration of person re-id challenges \cite{gong2014person}.
		{\bf Left}: great visual similarity among different people.
		{\bf Right}: large cross-view appearance variations of
		the same people, each person within a dotted box.
	}
	\vspace{-.6cm}
	\label{fig:challenge}
\end{figure}




In view of
the analysis above,
we formulate a novel view-specific person re-identification framework,
named \emph{\textbf{C}amera co\textbf{R}relation \textbf{A}ware \textbf{F}eature augmen\textbf{T}ation} (CRAFT),
capable of performing cross-view feature adaptation by measuring cross-view correlation
from visual data distribution
and carrying out adaptive feature augmentation to transform the original features
into a new augmented space.
Specifically, we quantify the underlying camera correlation in our framework by generalizing the conventional zero-padding, a non-parameterized feature augmentation mechanism, to a parameterized feature augmentation.
As a result, any two cameras can be modelled adaptively but not independently,
whereas the common information between camera views have already been quantified in the adaptive space.
Through this augmentation framework, view-generic learning algorithms can be readily generalized to induce view-specific sub-models whilst involving simultaneously view-generic discriminative modelling.
More concretely, we instantialize our CRAFT framework with Marginal Fisher Analysis (MFA) \cite{MFA}, leading to a re-id method instance called CRAFT-MFA.
We further introduce camera view discrepancy regularization in order to
append extra modelling capability for controlling the correlation degree between view-specific sub-models.
This regularization can be viewed as a complementary means of incorporating camera correlation modelling
on top of the proposed view-specific learning strategy.
Moreover, our CRAFT 
framework can be flexibly deployed for
re-id \zhu{across multiple} $(>2)$ camera views by jointly learning a unified model, 
which is largely under-studied in existing approaches.

Apart from cross-view discriminative learning,
we also investigate domain-generic 
(i.e., independent of target data or domain) person appearance representation,
  with the aim to make person features towards view invariant for
  facilitating the cross-view adaptation process \wsf{using} CRAFT.
In particular, we explore the potential of deep learning techniques
for person appearance description,
inspired by the great success of deep neural networks in
other related applications
like object recognition and detection \cite{Alexnet,decaf,RCNN}.
Differing significantly from existing deep learning based re-id approaches
\cite{Deepreid,ahmed2015improved,hu2015deep,Dropout_reid} that
typically learn directly from the small person re-id training data,
we instead utilize large auxiliary less-related image data.
This strategy allows to not only avoid the insufficient training data problem and address the limited scalability challenge in deep re-id models,
but also yield domain-generic person features
with more tolerance to view variations.


{
  The main {\bf contributions} of this work include}:
  {\bf (I)} {\wsff{We propose a
  camera correlation aware feature augmentation person re-id framework called CRAFT}. Our framework is able to generalize existing view-generic 
    person re-identification models to {perform view-specific learning}.
    A kernelization {formulation} is also presented.}
  %
%
%
%
%
%
%
  {\bf ({II})} We extend our CRAFT framework to
  {jointly learn view-specific feature transformations for person re-id across
a large network involving more than two cameras.
  Although this is a realistic scenario,
  how to build an effective unified re-id model for an entire camera network
  is still under-explored 
  in existing studies.}
  {\bf ({III})} We present a deep convolutional network based appearance feature extraction method in order to extract domain-generic and {more view invariant} person features.
  To our knowledge, this is the first attempt that {explores deep learning with large auxiliary non-person image data
  for constructing discriminative re-id features.}
For evaluating our method,
we conducted extensive comparisons between CRAFT and a variety of state-of-the-art
models on 
  VIPeR \cite{viper}, CUHK01 \cite{transferREID}, {CUHK03 \cite{Deepreid}}, 
  QMUL GRID \cite{loy2009multi}, and Market-1501 \cite{market} person re-id benchmarks.

\section{Related Work}
\label{sec:related_work}

\noindent {\bf {Distance metric} learning in person re-id. }
Supervised learning based methods \cite{RankSVM,zheng2011person,PRDC,LADF,KISSME,PCCA,kernelREIDECCV14,ensembleReID,wang2014person,XQDA,LiaoPSD,zheng2015partial,li2015multi,zhang2016learning,SCSP,TopPush,wang2016person,wang2016human}
dominate current person re-id research
by achieving state-of-the-art performance,
whilst a much fewer unsupervised re-id methods \cite{farenzena2010symmetric,eSDCocsvm,ISR,ElyorBMVC15,wang2016towards,ma2017person} have been proposed
with much inferior results yielded.
This is because large cross-camera variations in viewing conditions
may cause dramatic person appearance changes and arise great difficulties for
accurately matching identities.
Discriminative learning from re-id
training data is typically considered as a
necessary and effective strategy for reliable person re-id.
Notable re-id learning models include 
PRDC \cite{PRDC},
LADF \cite{LADF},
KISSME \cite{KISSME},
PCCA \cite{PCCA},
LFDA \cite{LFDA},
XQDA \cite{XQDA},
{PSD\cite{LiaoPSD}},
{Metric Ensemble \cite{ensembleReID}, DNS \cite{zhang2016learning}, {SCSP \cite{SCSP}}},
and so forth.

{All the above re-id methods are mainly designed for learning 
  the common view-generic discriminative knowledge,
but ignoring greatly the individual view-specific feature variation under each camera view.}
This limitation can be relaxed by the recent view-specific modelling strategy capable of learning
an individual matching function for each camera view. 
Typical methods of such kind include the CCA (canonical correlation
analysis) based approaches ROCCA~\cite{ROCCA}, refRdID~\cite{an2015person}, KCCA-based re-id~\cite{lisanti2014matching}, and CVDCA~\cite{chen2015CVDCA}.
However, they are less effective in extracting the shared discriminative information between different views, because these CCA-based methods~\cite{ROCCA,an2015person,lisanti2014matching} do not 
\xpami{directly/simultaneously} quantify the commonness and discrepancy between views during learning transformation, so that they cannot identify {accurately} 
what information can be shared between views.
While CVDCA~\cite{chen2015CVDCA} attempts to {quantify the inter-view discrepancy, it is {theoretically} restricted due to} the stringent Gaussian distribution assumption on person image data, which may yield sub-optimal modelling at the presence of {typically} complex/significant cross-view appearance changes.
%
%
%
View-specific modelling for person re-id remains under studied to a great extent.

In this work,
we present {a} different view-specific person re-id \wsf{framework},
characterized by a unique capability of generalizing view-generic distance metric learning methods to perform view-specific person re-id modelling, whilst still preserving their inherent learning strength. Moreover, our method can flexibly benefit many existing \wsf{distance metric/subspace-based} person re-id models for substantially improving re-id performance.


\vspace{0.1cm}
\noindent {\bf Feature representation in person re-id. }
Feature representation is another important issue for re-id.
Ideal person image features should be sufficiently
invariant against viewing condition changes and
generalized across different cameras/domains.
To this end, person re-id images are often represented
by hand-crafted appearance pattern based features, designed and computed according to human domain knowledge
\cite{wang2007shape,gray2008viewpoint,farenzena2010symmetric,cheng2011custom,bak2010person,colorInvariants,eSDCocsvm,ISR,hirzer2012relaxed,XQDA,GOG}.
These features are usually constituted by multiple different types of descriptors (e.g.,
color, texture, gradient, shape or edge) 
and greatly domain-generic 
(e.g., no need to learn from target labelled training data).
\xpami{Nowadays, deep feature learning for person re-id has attracted increasing attention \cite{Deepreid,yi2014deep,ahmed2015improved,FNNReID2016,chen2016deep,Wang_joint,Cheng_TCP,Dropout_reid,Gated_SCNN,varior2016siamese}.
These alternatives allow benefiting from the powerful modelling capacity of neural networks,
and are thus suitable for joint learning even given very heterogeneous training data \cite{Dropout_reid}.}
Often, they require a very large collection of labelled training data and can
easily suffer
from the model overfitting risk in realistic applications when data are not sufficiently provided 
\cite{ensembleReID,zhang2016learning}. 
Also, these existing deep re-id features are typically domain-specific.

%
In contrast,
our method exploits deep learning techniques
for automatically mining more diverse and view invariant 
appearance patterns
({\em versus} restricted hand-crafted ones)
from auxiliary less-relevant image data ({\em versus} using the sparse person re-id training data),
finally leading to more reliable person representation.
Furthermore, our deep feature is
largely domain-generic with no need for labelled target training data.
Therefore, our method possesses simultaneously the benefits of the two {conventional} feature
extraction paradigms above. 
The advantages of our proposed features over existing \wsf{popular} alternatives
are demonstrated in our evaluations (Section \ref{sec:eval_proposed} and Table \ref{tab:eval_feat}).

\vspace{0.1cm}
\noindent {\bf Domain adaptation.}
In a broader context, our cross-view adaptation \wsf{for re-id} is related to but different vitally from domain adaptation (DA) \cite{TCA,TFLDA,DAML,xiao2014KernelMatching,augmentedDALearn,gopalan2014unsupervised,DAzeroPadding,Peng_unsupervised}. Specifically, the aim of existing DA methods is to diminish the distribution discrepancy between the source and target domains, which is similar conceptually to our method. However, DA  models assume typically that the training and test classes (or persons) are overlapped, whilst our method copes with disjoint training and test person (class) sets with the objective to learn a discriminative model that is capable of generalizing to previously unseen classes (people). Therefore, \wsf{conventional} DA methods are less suitable for person re-id.

\zhu{
\vspace{0.1cm}
\noindent {\bf Feature augmentation.}
Our model formulation is also related to the zero padding technique \cite{muquet2002cyclic,wang2004multiple}
and \cycb{feature data augmentation methods} in
\cite{DAzeroPadding,augmentedDALearn}.
However, \cycb{ours is different substantially from these existing methods}.
Specifically, \cite{augmentedDALearn} is designed particularly for
a heterogeneous modelling problem (i.e., different feature representations
are used in distinct domains), whilst person re-id is typically
homogeneous and therefore not suitable.
More critically, beyond all these conventional methods,
our augmentation formulation uniquely considers the relation
between transformations of different camera views (even if more than two)
and embeds intrinsic camera correlation into the adaptive augmented feature space for facilitating cross-view feature adaptation and finally person identity association modelling.}
%



\section{Towards View Invariant Structured Person Representation}
\label{sec:feature}

\begin{figure*}
  \centering
  \includegraphics[width=1\textwidth]{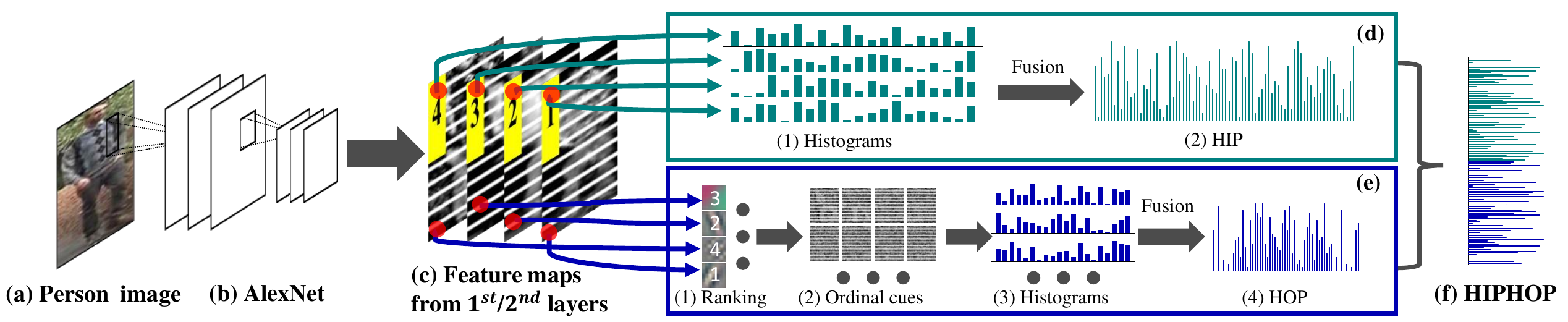}
  \vskip -.4cm
  \caption{\footnotesize
    Illustration of extracting the proposed HIPHOP feature.
    (a) A resized input person image with $227 \times 227$ pixel size.
    (b) Forward propagate the resized image through the whole AlexNet architecture.
    (c) Obtain the feature maps of the $1^\text{st}$ and $2^\text{nd}$ convolutional layers.
    (d) Compute the HIP descriptor by pooling activation intensity into histograms
    across different horizontal strips and feature maps.
    (e) Extract the HOP descriptor by ranking the localized activations
    and then pooling the top-$\kappa$ feature map indices over all horizontal strips
    and feature maps.
    (f) Construct the final HIPHOP feature by fusion.
       }
      \vspace{-.5cm}
  \label{feat:HIPHOP}
\end{figure*}

{We want to construct \wsf{more view change tolerant} person re-id features.
  To this end, we explore the potential of deep convolutional neural network,
  encouraged by its great generalization capability in related applications \cite{RCNN,visualizeCNN}.}
Typically, one has only sparse (usually hundreds {or thousands})
labelled re-id training {samples}
due to expensive data annotation.
{This leads to great challenges for deriving effective domain-generic features using deep models} \cite{Deepreid,ahmed2015improved,ding2015deep}.
We resolve the sparse training data problem
by learning the deep model with a large auxiliary image dataset,
rather than attempting to learn person discriminative features from the small re-id training data.
Our intuition is that, a {generic} person description
relies largely on the quality of atomic appearance patterns.
{Naturally, our method can be largely
  domain-generic in that appearance patterns learned from
  large scale data are likely to be more general}.
To this purpose,
we first exploit
the AlexNet\footnote{Other networks such as the VggNet \cite{simonyan2014very} and
  GoogLeNet \cite{szegedy2015going} architectures can be considered without any limitation.} \cite{Alexnet}
convolutional neural network (CNN) to learn from the ILSVRC 2012 training dataset
for obtaining diverse patterns.
Then, we design reliable descriptors to
characterize the appearance of a given person image.
We mainly leverage the lower convolutional (conv) layers,
unlike \cite{decaf} that utilizes the higher layers
(e.g., the $5^{\text{th}}$ conv layer,
$6^{\text{th}}/7^{\text{th}}$ fully connected (fc) layers).
The reasons are:
(1) Higher layers can be more likely to
be task-specific
(e.g., sensitive to general object categories rather than
person identity in our case), and
may have worse transferability
than lower ones \cite{decaf,visualizeCNN}
(see our evaluations in Table \ref{tab:eval_feat}).
In contrast, lower conv layers correspond to
low-level visual features such as
color, edge and elementary texture patterns \cite{visualizeCNN}\wsf{, }
and naturally have better generality across different tasks.
(2) Features from higher layers are possibly contaminated by dramatic variations in human pose {or}
background clutter due to their large receptive fields
and thus not sufficiently localized for person re-id.

We present two person descriptors based on feature maps of
the first two conv layers\footnote{
  More conv layers can be utilized similarly but at the cost
  of increasing the feature dimension size.}, as detailed below.

\vspace{0.1cm}

\noindent {\bf Structured person descriptor. }
The feature data from convnet layers are highly structured,
e.g., they are organized in form of multiple 2-D feature maps.
Formally, we denote by $\bm{F}_c \in \mathbb{R}^{h_c\times w_c\times m_c}$
the feature maps of the $c$-th ($c \in \{1, 2\}$) layer,
with $h_c$/$w_c$ the height/width of feature maps,
and $m_c$ the number of conv filters.
  In our case,
$h_1 = w_1 = 55, m_1 = 96$; and $h_2=w_2=27, m_2 = 256$.
For the $c$-th layer,
$\bm{F}_c(i,j,\kappa)$ represents the activation of the $\kappa$-th filter on
the image patch \wsf{centred} at the pixel $(i,j)$.
Given the great variations in person pose, we further divide each
feature map into horizontal strips with a fixed height $h_s$
for encoding spatial structure information and enhancing pose invariance,
similar to ELF \cite{gray2008viewpoint} and WHOS \cite{ISR}.
We empirically set $h_s = 5$ in our evaluation
for preserving sufficient spatial structure.
We then extract the intensity histogram (with bin size 16)
from each strip for every feature map.
The concatenation of all these strip based histograms
forms the Histogram of Intensity Pattern ({\bf HIP}) descriptor,
with feature dimension
$37376=16(\text{bins})\times11(\text{strips})\times96(m_1) +
  16(\text{bins})\times5(\text{strips})\times 256(m_2)$.
As such, HIP is inherently structured, containing multiple components
with different degrees of pattern abstractness.

\vspace{0.1cm}

\noindent {\bf View-invariant person descriptor. }
The proposed HIP descriptor encodes completely
the activation information of all feature maps,
regardless their relative saliency and noise degree.
This ensures pattern completeness, but 
being potentially sensitive to cross-view covariates
such as human pose changes and background discrepancy.
To mitigate this issue,
we propose to selectively use these feature maps 
for introducing further view-invariance capability.
This is realized
by incorporating the activation ordinal information \cite{ordinalPalm,SinhaPHD},
yielding another descriptor called
Histogram of Ordinal Pattern ({\bf HOP}).
\xpami{We similarly encode the spatial structure information by the same horizontal decomposition as HIP.}

Specifically,
we rank all activations $\{\bm{F}_c(i,j,\kappa)\}_{\kappa=1}^{m_c}$
{in descendant order}
and get the top-$\kappa$ feature map indices, denoted as
\begin{equation}\label{eq:hop:rank}
\bm{p}_c(i,j) =  [v_1,v_2, \cdots, v_\kappa],
\end{equation}
where $v_i$ is the index of the $i$-th feature map in the ranked activation list.
We fix $\kappa = 20$ in our experiments.
By repeating the same ranking process for each image patch,
we obtain a new representation $\bm{P}_c \in \mathbb{R}^{h_c\times w_c\times \kappa}$
with elements $\bm{p}_c(i,j)$.
Since $\bm{P}_c$ can be considered as another set of feature maps,
we utilize a similar pooling way as HIP to construct its histogram-like representation,
but with bin size $m_c$ for the $c$-th layer.
Therefore, the feature dimension of HOP is
$46720=96(\text{bins}, m_1)\times11(\text{strips})\times20(\kappa) +
256(\text{bins}, m_2)\times5(\text{strips})\times20(\kappa)$.
Together with HIP, we call our final fused person feature as {\bf HIPHOP},
with the total dimension $84096=46720+37376$.

\vspace{0.1cm}
\noindent {\bf Feature extraction overview. }
We depict the main steps of constructing
our HIPHOP feature. 
First, we resize a given image into the size of $227$$\times$$227$,
as required by AlexNet (Figure \ref{feat:HIPHOP}(a)).
Second, we forward propagate the
resized image through the AlexNet (Figure \ref{feat:HIPHOP}(b)).
Third, we obtain the feature maps from the $1^\text{st}$ and $2^\text{nd}$ conv layers
(Figure \ref{feat:HIPHOP}(c)).
Fourth, we compute the HIP (Figure \ref{feat:HIPHOP}(d)) and HOP
(Figure \ref{feat:HIPHOP}(e)) descriptors.
Finally, we composite the HIPHOP feature for the given person image
by vector concatenation 
(Figure \ref{feat:HIPHOP}(f)).
For approximately suppressing background noise, we impose an Epanechnikov kernel \cite{ISR} as weight on each activation map before computing histogram.

\section{Camera Correlation Aware Feature Augmentation for Re-id}

We formulate a novel {view-specific} 
person re-id framework,
namely \textbf{C}amera co\textbf{R}relation \textbf{A}ware \textbf{F}eature augmen\textbf{T}ation (CRAFT), to adapt the original image features into another view adaptive space,
where many {view-generic} methods can be readily deployed
for achieving view-specific discrimination modelling.
\wsf{In the following, we \wsff{formulate} re-id
  as a \zhu{feature augmentation problem,} 
  and then present our \zhu{CRAFT framework.} \wsff{We first discuss the re-id under two non-overlapping camera views and later generalize our model under multiple \zhux{(more than two)} camera views.}
  }

\subsection{Re-Id Model Learning Under Feature Augmentation} \label{sec:feat_aug}

\wsff{Given \zhux{image} data from two non-overlapping camera views, namely camera $a$ and camera $b$,} we reformulate {person re-id in \wsf{a} feature augmentation framework.}
Feature augmentation has been exploited in the domain adaptation problem.
For example, Daum{\'e} III \cite{DAzeroPadding}
proposed the feature mapping functions
$\rho^s (\bm{x}) = [\bm{x}^\top, \bm{x}^\top, (\bm{0}_{d})^\top]^\top$ (for the source domain) and
$\rho^t (\bm{x}) = [\bm{x}^\top, (\bm{0}_{d})^\top, \bm{x}^\top]^\top$ (for the target domain)
for homogeneous domain adaptation, with
$\bm{x} \in \mathbb{R}^d$ denoting the sample feature,
$\bm{0}_{d}$ the $d$ column vector of all zeros,
$d$ the feature dimension, and the superscript $^\top$ the transpose of a vector or a matrix. \cyc{This can be viewed as incorporating the original feature into an augmented space for enhancing the similarities between data from the same domain
and thus increasing the impact of same-domain (or same-camera) data.
For person re-id, this should be unnecessary given its cross-camera matching nature.}
Without original features in augmentation,
they are resorted to zero padding,
a technique widely exploited in signal transmission \cite{muquet2002cyclic,wang2004multiple}.
%
%

\vspace{0.1cm}
\noindent {\bf Zero padding. }
Formally, the zero padding augmentation can be formulated as:
\begin{equation} \label{eq:aug:zeroPad}
\begin{array}{ll}
\tilde{\bm{X}}_{\text{zp}}^a =
\begin{bmatrix}
\bm{I}_{d \times d} \\[0.3em]
\bm{O}_{d \times d}
\end{bmatrix}
\bm{X}^a, \;\;\;\;\;\;\;\;
\tilde{\bm{X}}_{\text{zp}}^b =
\begin{bmatrix}
\bm{O}_{d \times d} \\[0.3em]
\bm{I}_{d \times d}
\end{bmatrix}
\bm{X}^b
\end{array},
\end{equation}
where ${\bm{X}}^a = [\bm{x}_1^{a}, \dots, \bm{x}_{n_a}^{a}] \in \mathbb{R}^{d \times n_a}$ 
and
$\bm{X}^b = [\bm{x}_1^{b}, \dots, \bm{x}_{n_b}^{b}] \in \mathbb{R}^{d \times n_b}$
represent the column-wise image feature matrix from camera $a$ and $b$;
$\tilde{\bm{X}}_{\text{zp}}^a = [\tilde{\bm{x}}_{\text{zp},1}^{a}, \dots, \tilde{\bm{x}}_{\text{zp},n_a}^{a}] \in \mathbb{R}^{2d \times n_a}$ and
$\tilde{\bm{X}}_{\text{zp}}^b = [\tilde{\bm{x}}_{\text{zp},1}^{b}, \dots, \tilde{\bm{x}}_{\text{zp},n_b}^{b}] \in \mathbb{R}^{2d \times n_b}$ refer to augmented feature matrices;
\cycb{
$n_a$ and $n_b$ are the training sample numbers of camera $a$ and camera $b$, respectively;
$\bm{I}_{d \times d}$ and $\bm{O}_{d \times d}$ denote the $d \times d$ identity matrix and zero matrix, respectively.}

\vspace{0.2cm}
\noindent {\bf {Re-id {reformulation}}. }
The augmented features \cycb{$\tilde{\bm{X}}_{\text{zp}}^\phi$ ($\phi\in\{a,b\}$)} can be incorporated into different existing view-generic distance \wsf{metric} or subspace learning algorithms.
Without loss of generality, we take the subspace learning as example in the follows.
Specifically, the aim of discriminative learning is to estimate the optimal projections $\hat{\bm{W}} \in \mathbb{R}^{2d \times m}$ (with $m$ the subspace dimension)
such that after projection \cycb{$\bm{z}^\phi = \hat{\bm{W}}^\top \tilde{\bm{x}}_{\text{zp}}^\phi$, where $\phi \in \{a,b\}$ and $\tilde{\bm{x}}_{zp}^\phi$ is an augmented feature defined in Eqn. \eqref{eq:aug:zeroPad},}
one can effectively discriminate between different identities
by Euclidean distance.
Generally, 
the objective function can be written as
\begin{equation}\label{eq:gen_obj}
  \hat{\bm{W}} = \min_{\bm{W}}
  {f}_{\text{obj}} \cycc{(}
  {\bm{W}}^\top\tilde{\bm{X}}_{\text{zp}} \cycc{),}
\end{equation}
\cycc{where $\tilde{\bm{X}}_{\text{zp}} = [\tilde{\bm{X}}_{\text{zp}}^a, \tilde{\bm{X}}_{\text{zp}}^b] = [\tilde{\bm{x}}_{\text{zp},1}, \dots, \tilde{\bm{x}}_{\text{zp},n}]$\cycd{, $n=n_a+n_b$. $\tilde{\bm{X}}_{\text{zp}}$ is the combined feature data matrix from camera $a$ and camera $b$.}}

Clearly, $\hat{\bm{W}}$ can be decomposed into two parts as:
\begin{equation}
\hat{\bm{W}} = [(\hat{\bm{W}}^a)^\top, (\hat{\bm{W}}^b)^\top]^\top ,
\label{eq:submodels}
\end{equation}
with∫
$\hat{\bm{W}}^a \in \mathbb{R}^{d \times m}$ and
$\hat{\bm{W}}^b \in \mathbb{R}^{d \times m}$
corresponding to the respective projections (or sub-models) for camera $a$ and $b$.
This is due to the zero padding based feature augmentation
(Eqn. (\ref{eq:aug:zeroPad})):
\begin{equation}\label{eqn:prediction_zero}
\begin{array}{ll}
\hat{\bm{W}}^\top \tilde{\bm{X}}^a_{\text{zp}} = (\hat{\bm{W}}^a)^\top \bm{X}^a, \\ [0.1cm]
\hat{\bm{W}}^\top \tilde{\bm{X}}^b_{\text{zp}} = (\hat{\bm{W}}^b)^\top \bm{X}^b.
\end{array}
\end{equation}
Clearly, 
zero padding allows {\em view-generic} methods to 
\cycd{simultaneously} learn two {\em view-specific} sub-models,
i.e., $\hat{\bm{W}}^a$ for view $a$ and $\hat{\bm{W}}^b$ for view $b$,
and realize {\em view-specific} modelling \cite{ROCCA,an2015person,lisanti2014matching},
likely better aligning cross-view image data distribution \cite{DAzeroPadding,chen2015CVDCA}.


\begin{figure}
  \centering
  \subfigure[\footnotesize Original 1-D feature space]{
    \includegraphics[height=0.37\linewidth]{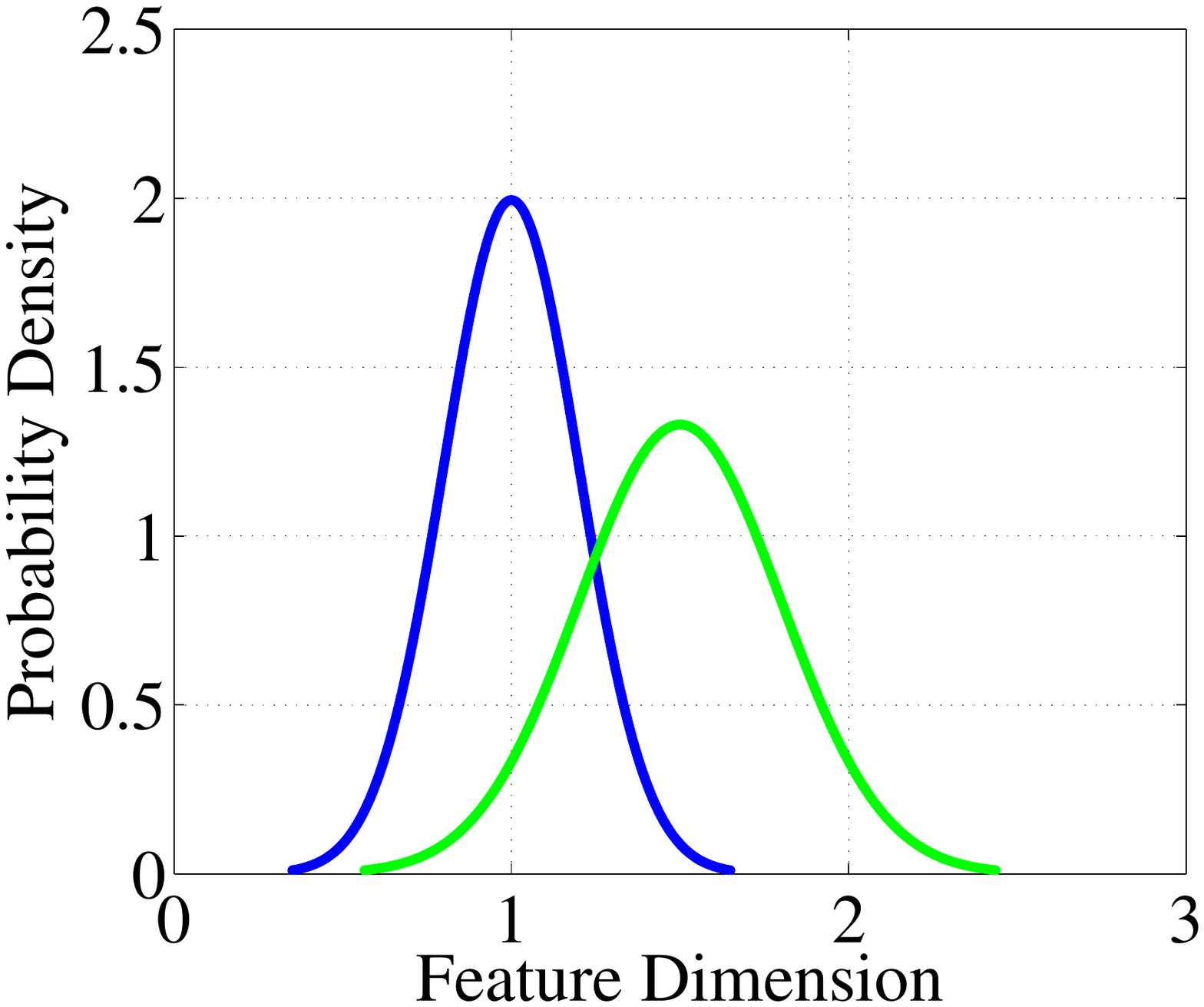}
  }
  \hfill
  \subfigure[\footnotesize Augmented 2-D feature space]{
    \includegraphics[height=0.37\linewidth]{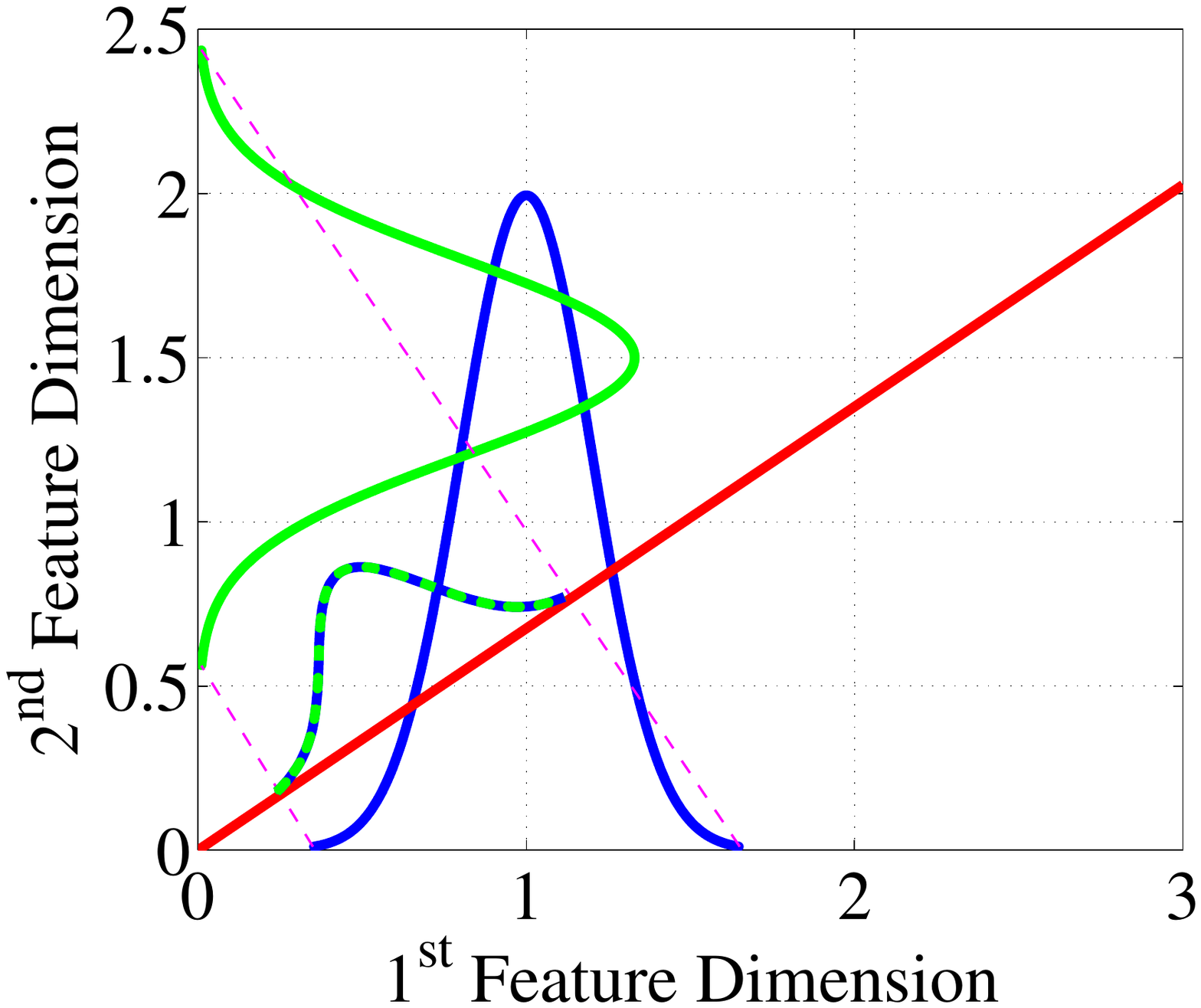}
  }
  \vskip -.3cm
  \caption{\footnotesize
  	\xpami{
    An illustration of zero padding based feature augmentation. 
    (a) The data distribution in the original feature space 
    from 
    camera view $a$ (the blue curve) and $b$ (the green curve). 
    (b) The augmented feature space by zero padding. 
    The dashed blue and green curves represent the projected features
    with respect to the projection basis indicated by the solid red line.
	The two dashed lines imply feature projection operation. 
  {Note that the probability density axis is not plotted in (b) for demonstration simplicity}.}
  }
  \vspace{-.5cm}
  \label{feat:ZP}
\end{figure}

\xpami{For better understanding, we take an example in 1-D feature space. 
  Often, re-id feature data distributions from different camera views are misaligned due to
  the inter-camera viewing condition discrepancy (Figure \ref{feat:ZP}(a)). 
  With labelled training data, the transformation learned in Eqn. \eqref{eq:gen_obj}
  aims to search for an optimized 
  projection in the augmented feature space (the red line in Figure \ref{feat:ZP}(b)) such that 
  cross-view data distributions are aligned and thus good for matching images of the same person across camera views.
%
Clearly, the zero-padding treats each camera view independently by 
optimizing two separated view-specific sub-models and therefore 
allows to better quantify the feature distortion of either camera view.
Nonetheless, as compared to a single view-generic model, this doubled modelling space 
may unfavourably loosen inter-camera inherent correlation (e.g., the ``same'' person with ``different'' appearance in the images captured by two cameras with distinct viewing conditions).
This may in turn make the model optimization less effective 
in capturing appearance variation across views
and extracting shared view-generic discriminative cues.} 

To overcome the above limitation, we design particularly the camera correlation aware feature augmentation, 
which allows for adaptively incorporating
the common information between camera views into the augmented space
whilst retaining the capability of well modelling 
the feature distortion of individual camera views.

\subsection{Camera Correlation Aware Feature Augmentation}
\label{sec:feat_transformation}

The proposed feature augmentation method is performed in two steps:
(I) We quantify automatically the commonness degree between different camera views.
(II) We exploit the estimated camera commonness information for
adaptive feature augmentation.

\vspace{0.1cm}
\noindent {\bf (I) Quantifying camera commonness by correlation. }
We propose exploiting the correlation in image data distributions
for camera commonness quantification.
Considering that many different images may be generated by any camera,
we represent a camera by a set of images captured by itself,
e.g., {\em a set of feature vectors}.
We exploit the available 
training images captured by both cameras
for obtaining more reliable commonness measure.

Specifically, given image features
$\bm{X}^a$ and $\bm{X}^b$ for camera $a$ and $b${, respectively,}
we adopt the principle angles \cite{golub2012matrix}
to measure the correlation between the two views.
In particular, first,
we obtain the linear subspace representations by the principle component analysis,
$\bm{G}^a \in \mathbb{R}^{n_a \times \cycd{r}}$ for $\bm{X}^a$ and
$\bm{G}^b \in \mathbb{R}^{n_b \times \cycd{r}}$ for $\bm{X}^b$,
with $r$ the dominant component number.
In our experiments, we empirically set $\cycd{r}=100$. 
Either $\bm{G}^a$ or $\bm{G}^b$ can then be seen
as a data point on the Grassmann manifold --
a set of fixed-dimensional linear subspaces of Euclidean space.
Second, we measure the similarity between the two manifold points
with their principle angles
($0 \leq\theta_1 \leq \cdots \leq \theta_k \leq \cdots \leq \theta_{\cycd{r}} \leq \frac{\pi}{2}$)
defined as:
\begin{equation}
\begin{array}{ll}
\cos(\theta_k) = \max \limits_{\bm{q}_j \in \text{span}(\bm{G}^a)}
\max \limits_{\bm{v}_k \in \text{span}(\bm{G}^b)} \bm{q}_k^\top \bm{v}_k ,\\ [0.3em]
\text{s.t. } \quad
\bm{q}_k^\top \bm{q}_k = 1, \quad
\bm{v}_k^\top \bm{v}_k = 1, \\ [0.3em]
\quad \quad \;\;
\bm{q}_k^\top \bm{q}_i = 0, \quad
\bm{v}_k^\top \bm{v}_i = 0, \quad i \in [0, k-1] ,
\end{array}
\end{equation}
where $\text{span}(\cdot)$ denotes the subspace spanned
by the column vectors of a matrix.
The intuition is that,
principle angles have good geometry interpretation
(e.g., related to the manifold geodesic distance \cite{wong1967differential,GDA})
and their cosines $\cos(\theta_k)$ are known as {\em canonical correlations}.
Finally, we estimate the camera correlation (or commonness degree) $\omega$ as:
\begin{equation}\label{eq:estimate_r}
\omega = \frac{1}{\cycd{r}} \sum_{k=1}^{\cycd{r}} \cos(\theta_k),
\end{equation}
with 
$\cos(\theta_k)$ computed 
by
Singular Value Decomposition:
\begin{equation}\label{PrincAngleOfViews}
(\bm{G}^a)^\top \bm{G}^b = \cycd{\bm{Q}} \cos(\Theta) \bm{V}^\top,
\end{equation}
where
$\cos(\Theta) = \text{diag}(\cos(\theta_1), \cos(\theta_2), \cdots \cos(\theta_{\cycd{r}}))$,
\cycd{$\bm{Q} = [\bm{q}_1, \bm{q}_2, \cdots, \bm{q}_{\cycd{r}}]$},
and $\bm{V} = [\bm{v}_1, \bm{v}_2, \cdots, \bm{v}_{\cycd{r}}]$.



\vspace{0.1cm}
\noindent {\bf (II) Adaptive feature augmentation. }
Once obtaining the camera correlation $\omega$,
we want to incorporate it into feature augmentation.
To achieve this, we generalize the zero padding (Eqn. \eqref{eq:aug:zeroPad}) to:
\begin{equation} \label{eq:general_transformation}
\begin{array}{ll}
\tilde{\bm{X}}_{\text{craft}}^a =
\begin{bmatrix}
\bm{R} \\[0.3em]
\bm{M}
\end{bmatrix}
\bm{X}^a, \;\;\;\;\;\;\;\;
\tilde{\bm{X}}_{\text{craft}}^b =
\begin{bmatrix}
\bm{M} \\[0.3em]
\bm{R}
\end{bmatrix}
\bm{X}^b
\end{array},
\end{equation}
where $\bm{R}$ and $\bm{M}$ refer to the $d \times d$ augmentation matrices.
So, zero padding is a special case of the proposed feature augmentation (Eqn. \eqref{eq:general_transformation}) where
$\bm{R} = \bm{I}_{d \times d}$ and
$\bm{M} = \bm{O}_{d \times d}$.

\wsf{With some view-generic discriminative learning algorithm,
we can learn an optimal model
$\bm{W} = [(\bm{W}^a)^\top, (\bm{W}^b)^\top]^\top$
in our augmented space. 
Then, feature mapping functions can be written:
\begin{equation}\label{eqn:prediction_general}
\small
\begin{array}{ll}
f_a (\tilde{\bm{X}}_{\text{craft}}^a)
= \bm{W}^\top \tilde{\bm{X}}_{\text{craft}}^a
= (\bm{R}^\top \bm{W}^a+ \bm{M}^\top \bm{W}^b)^\top  \bm{X}^a, \\ [0.1cm]
f_b (\tilde{\bm{X}}_{\text{craft}}^b)
= \bm{W}^\top \tilde{\bm{X}}_{\text{craft}}^b
= (\bm{M}^\top \bm{W}^a + \bm{R}^\top \bm{W}^b)^\top  \bm{X}^b.
\end{array}
\end{equation}
\zhu{Compared to zero padding (Eqn. (\ref{eqn:prediction_zero})),}
  it is clear that \zhu{the feature transformation for
  each camera view is not treated independently in our adaptive feature augmentation (Eqn. \eqref{eqn:prediction_general}). Instead,}
  the transformations of all camera views are intrinsically \zhu{correlated} and meanwhile \zhu{being} view-specific.}

\wsf{However}, it is non-trivial 
to estimate automatically the augmentation matrices $\bm{R}$ and $\bm{M}$
with the estimated camera correlation information accommodated
for enabling more accurate cross-view discriminative analysis (Sec. \ref{sec:reg}).
This is because
a large number of ($2d^2$) parameters are required to be learned
given the typically high feature dimensionality $d$ (e.g., tens of thousands)
but only a small number of (e.g., hundreds) training data available.
%
Instead of directly learning from the training data,
we propose to properly design $\bm{R}$ and $\bm{M}$ for
overcoming this problem as:
\begin{equation} \label{eq:def:myPQ}
\bm{R} = \frac{2 - \omega}{\varpi} \bm{I}_{d \times d},
\;\;\;\;\;\;\;\;
\bm{M} = \frac{\omega}{\varpi} \bm{I}_{d \times d} ,
\end{equation}
where \cycb{$\omega$ is  the camera correlation defined in Eqn. \eqref{eq:estimate_r} and }
$\varpi = \sqrt{(2-\omega)^2 + \omega^2}$ is the normalization term.
In this way, camera correlation is directly embedded into the feature augmentation process.
Specifically, when $\omega = 0$,
which means the two camera views are totally uncorrelated with no common property,
we have $\bm{M}= \bm{O}_{d \times d}$ and $\bm{R} = \bm{I}_{d \times d}$,
and our feature augmentation degrades to zero padding.
When $\omega = 1$,
which means the largest camera correlation,
{we have} $\bm{R} = \bm{M}$
thus potentially similar view-specific sub-models,
i.e., strongly correlated each other.
In other words, $\bm{M}$ represents the shared degree across camera views
whilst $\bm{R}$ stands for view specificity strength,
with their balance controlled by the inherent camera correlation.

\wsf{\zhu{By} using Eqn. (\ref{eq:def:myPQ}), the view-specific feature mapping functions in Eqn. (\ref{eqn:prediction_general})
  \zhu{can be expressed as:}}
\begin{equation}\label{eqn:prediction}
\begin{array}{ll}
f_a (\tilde{\bm{X}}_{\text{craft}}^a)
& = \underbrace{\frac{2 - \omega}{\varpi}(\bm{W}^a)^\top}_{\text{specificity}} \bm{X}^a + \underbrace{\frac{\omega}{\varpi}(\bm{W}^b)^\top}_{\text{adaptiveness}} \bm{X}^a,
\\ [0.3cm]
f_b (\tilde{\bm{X}}_{\text{craft}}^b)
& = \underbrace{\frac{\omega}{\varpi}(\bm{W}^a)^\top}_{\text{adaptiveness}} \bm{X}^b +
\underbrace{\frac{2-\omega}{\varpi} (\bm{W}^b)^\top}_{\text{specificity}} \bm{X}^b.
\end{array}
\end{equation}
Obviously, the mapped discriminative features for each camera view depend on
its respective sub-model
(weighted by $\frac{2-\omega}{\varpi}$, corresponding to view-specific modelling)
as well as the other sub-model
(weighted by $\frac{\omega}{\varpi}$, corresponding to view-generic modelling).
As such, \wsf{
  \zhu{our proposed} transformations realize}
the joint learning of
both view-generic and view-specific discriminative information.
We call this {\em cross-view adaptive} modelling



\vspace{0.2cm}

\noindent \textbf{Model formulation analysis} -
We examine the proposed formulation by analyzing the theoretical
connection among model parameters $\{\bm{R}, \bm{M}, \bm{W}\}$.
Note that in our whole modelling,
the augmentation matrices $\bm{R}$ and $\bm{M}$ (Eqn. \eqref{eq:def:myPQ}) are appropriately designed
with the aim for embedding the underlying camera correlation into a new feature space,
whereas $\bm{W}$ (Eqn. \eqref{eq:gen_obj}) is automatically learned from the training data.
Next, we demonstrate that learning $\bm{W}$ alone is sufficient to obtain the optimal solution.

Formally, we denote the optimal augmentation matrices:
\begin{equation}\label{equation:r_opt}
\bm{R}^{\text{opt}} = \bm{R} + \nabla \bm{R}, \quad 
\bm{M}^{\text{opt}} = \bm{M} + \nabla \bm{M}, 
\end{equation}
with $\nabla \bm{R}$ the difference (e.g., the part learned from the training data by
some ideal algorithm) between
our designed $\bm{R}$ and the assumed optimal one $\bm{R}^{\text{opt}}$ 
(similarly for $\nabla \bm{M}$).
The multiplication operation between $\bm{M}$ (or $\bm{R}$) and $\bm{W}$
in Eqn. \eqref{eqn:prediction} suggests that
\begin{equation}
\left\{
\begin{array}{ll}
(\bm{R} + \nabla \bm{R})^{\top}\bm{W}^a   + (\bm{M} + \nabla \bm{M})^{\top}\bm{W}^b  \\[0.05cm]
~~~= \bm{R}^\top (\bm{W}^a + \nabla \bm{W}^a)  + \bm{M}^\top(\bm{W}^b + \nabla \bm{W}^b)  \\ [0.15cm]
(\bm{M} + \nabla \bm{M})^{\top} \bm{W}^a + (\bm{R} + \nabla \bm{R})^{\top} \bm{W}^b \\[0.05cm]
~~~=\bm{M}^{\top}  (\bm{W}^a + \nabla \bm{W}^a) + \bm{R}^{\top} (\bm{W}^b + \nabla \bm{W}^b)
\end{array}
\right. ,
\end{equation}
where $\nabla \bm{W}^a$ and $\nabla \bm{W}^b$ are:
\begin{equation}\label{eq:map_change_tranf1}
\footnotesize
\left(
\begin{array}{c}
\nabla \bm{W}^a \\
\nabla \bm{W}^b \\
\end{array}
\right)
=
\left(
\begin{array}{cc}
\bm{R}^\top & \bm{M}^\top \\
\bm{M}^\top & \bm{R}^\top \\
\end{array}
\right)^{-1}
\left(
\begin{array}{cc}
\nabla \bm{R} & \nabla \bm{M} \\
\nabla \bm{M} & \nabla \bm{R} \\
\end{array}
\right)^\top
\left(
\begin{array}{c}
\bm{W}^a \\
\bm{W}^b \\
\end{array}
\right).
\end{equation}
Eqn. \eqref{eq:map_change_tranf1} indicates that the learnable parts $\nabla \bm{R}$
and $\nabla \bm{M}$ can be equivalently obtained in the process of optimizing
$\bm{W}$.
This suggests no necessity of directly learning $\bm{R}$ and $\bm{M}$ from the training data as long as
$\bm{W}$ is inferred through some optimization procedure.
{Hence, {deriving} $\bm{R}$ and $\bm{M}$
{as in Eqn. {\eqref{eq:def:myPQ}} should not degrade the effectiveness of our whole model, but instead making our entire formulation design more tractable and more elegant with different components fulfilling a specific function. }}


\subsection{Camera View Discrepancy Regularization}
\label{sec:reg}

\wsf{As aforementioned in Eqn. (\ref{eqn:prediction_general}), our transformations of all camera views are not independent \zhu{to each other}.
  Although these transformations are view-specific, they are mutually \zhu{correlated} in practice because they quantify the same group of people for association between camera views}. %
View-specific modelling (Eqns. \eqref{eq:submodels} and \eqref{eqn:prediction})
allows naturally {for} {regularizing} the mutual relation between 
different sub-models,
potentially \wsf{incorporating complementary correlation modelling between camera views in} \zhu{addition to Eqn. (\ref{eqn:prediction_general}).}
To achieve this, we enforce \wsf{the} constraint on sub-models 
by introducing a Camera View Discrepancy (CVD) regularization as:
\begin{equation}
\gamma_{\text{cvd}} = ||\bm{W}^a - \bm{W}^b||^2
\label{eqn:model_reg},
\end{equation}
\zhu{Moreover}, this CVD constraint can be combined well with the common ridge regularization as:
\begin{equation}
\begin{array} {ll}
  \gamma
  &= ||\bm{W}^a - \bm{W}^b||^2 + {\eta_{\text{ridge}}} \text{tr} (\bm{W}^\top\bm{W}) \\ [0.5em]
  &= \text{tr} (\bm{W}^\top
  \begin{bmatrix}
  \bm{I} & -\bm{I} \\[0.1em]
  -\bm{I} & \bm{I}
  \end{bmatrix}
  \bm{W}) + {\eta_\text{ridge}} \text{tr} ( \bm{W}^\top\bm{W}) \\ [1em]
  &=  
  {(1+\eta_\text{ridge})} \text{tr} (\bm{W}^\top
  \bm{C}  
  \bm{W}), \\
\end{array}
  \label{eqn:reg}
\end{equation}
where
\begin{equation}\label{eqn:C_pair}
  \bm{C} = \begin{bmatrix}
\bm{I} & -\beta \bm{I} \\ \nonumber
-\beta \bm{I} & \bm{I}
\end{bmatrix}, \quad
\beta = \frac{1}{1+{\eta_\text{ridge}}}.
\end{equation}
\cycb{$\text{tr} (\cdot)$ denotes the trace operation}, and ${\eta_\text{ridge}} > 0$ is a tradeoff parameter for balancing the
two terms.
The regularization $\gamma$ can be readily incorporated
into existing learning methods \cite{MFA,martinez2001pca,LFDA,KISSME,XQDA}
for possibly obtaining better model generalization.
Specifically, we define an enriched objective function on top of Eqn. \eqref{eq:gen_obj} as:
\begin{equation}
\label{general_obj}
\hat{\bm{W}}
= {\text{arg}} \min \limits_{\bm{W}} {f}_{\text{obj}}
(\bm{W}^\top \tilde{\bm{X}}_{\text{craft}})
+
\lambda \text{tr}( \bm{W}^\top \bm{C} \bm{W} ),
\end{equation}
where $\lambda$ controls the influence of $\gamma$.
Next, we derive the process for solving Eqn. \eqref{general_obj}.

\vspace{0.1cm}
\noindent {\bf View discrepancy regularized transformation. }
Since $\beta = \frac{1}{1+\eta_\text{ridge}} < 1$,
the matrix $\bm{C}$ is of \wsff{positive-definite}.
Therefore, $\bm{C}$ can be factorized into the form of
\begin{equation}
\bm{C} = \bm{P} \bm{\Lambda} \bm{P}^\top,
\label{eqn:factorize}
\end{equation}
with
$\bm{\Lambda}$ a diagonal matrix and
$\bm{P} \bm{P}^\top = \bm{P}^\top \bm{P} = \bm{I}$.
So, $\bm{P}^\top \bm{C} \bm{P}=\bm{\Lambda}$.
By defining
\begin{equation}
\bm{W} = \bm{P} \bm{\Lambda}^{-\frac{1}{2}} \bm{H},
\label{eqn:W2H}
\end{equation}
we have
\begin{equation}
\bm{W}^\top \bm{C} \bm{W} = \bm{H}^\top \bm{H}.
\end{equation}
Thus, Eqn. \eqref{general_obj} can be transformed equivalently to: 
\begin{equation}
\label{general_obj_2}
\hat{\bm{H}}
= {\text{arg}} \min \limits_{\bm{H}}
f_{\text{obj}}
  (\bm{H}^\top\bm{\Lambda}^{-\frac{1}{2}}\bm{P}^\top \tilde{\bm{X}}_{\text{craft}})
+ \lambda \text{tr}(\bm{H}^\top\bm{H}).
\end{equation}
\cycc{We define the transformed data matrix from all views}
\begin{equation}
\ddot{\bm{X}}_{\text{craft}}
= \bm{\Lambda}^{-\frac{1}{2}} \bm{P}^\top \tilde{\bm{X}}_{\text{craft}}
=  \wsff{[ \ddot{\bm{x}}_1,\cdots,\ddot{\bm{x}}_{n}]
  },
\label{eqn:RDFS}
\end{equation}
which we call {\em view discrepancy regularized transformation}.
%
So, Eqn. \eqref{general_obj_2} can be simplified as:
\begin{equation}
\label{general_obj_3}
\begin{array} {ll}
\hat{\bm{H}}
& = {\text{arg}} \min \limits_{\bm{H}}
f_{\text{obj}}(\bm{H}^\top  \ddot{\bm{X}}_{\text{craft}})
+ \lambda \text{tr}(\bm{H}^\top \bm{H}).
\end{array}
\end{equation}

\vspace{0.1cm}
\noindent {\bf Optimization. }
Typically, the same optimization algorithm as the adopted view-generic discriminative learning method
can be exploited to solve the optimization problem.
For providing a complete picture, we will present a case study
with a specific discriminative learning method incorporated into the proposed CRAFT framework.

\begin{algorithm} [t] \footnotesize 
  \caption{\footnotesize 
    Learning CRAFT-MFA
    }
  \label{Alg:CRAFT}
  \SetAlgoLined
  \KwIn
  {
    Training data $\bm{X}^a$ and $\bm{X}^b$ with identity labels;  
  }
  \vspace{.1cm} 
  \KwOut
  {
    Augmentation matrices $\bm{R}$ and $\bm{M}$,
    projection matrix $\hat{\bm{W}}$; 
  }
  
  \vspace{0.1cm}
  \textbf{(I) Camera correlation aware feature augmentation} (Section \ref{sec:feat_transformation}) \\ 
  - Estimate the camera correlation $\omega$
  (Eqn. \eqref{eq:estimate_r}); \\ 
  - Compute augmentation matrix $\bm{R}$ and $\bm{M}$
  (Eqn. \eqref{eq:def:myPQ}); \\ 
  - Transform original features \zhuxt{$\bm{X}$ into $\tilde{\bm{X}}_{\text{craft}}$}
  (Eqn. \eqref{eq:general_transformation}); \\

  \vspace{0.1cm}
  \textbf{(II) Camera view discrepancy regularization} (Section \ref{sec:reg}) \\ 
  - Obtain the camera view discrepancy regularization
  (Eqn. \eqref{eqn:model_reg}); \\ 
  - Get the fused regularization; (Eqn. \eqref{eqn:reg}); \\ 
  
  - Decompose $\bm{C}$ into $\bm{P}$ and $\bm{\Lambda}$
    (Eqn. \eqref{eqn:factorize}); \\ 
  - Transform \zhuxt{$\tilde{\bm{X}}_{\text{craft}}$ into $\ddot{\bm{X}}_{\text{craft}}$}
  (Eqn. \eqref{eqn:RDFS}); \\ 
  
  \vspace{0.1cm}
  \textbf{(III) Optimization} \\ 
  - Obtain $\hat{\bm{H}}$ with {the MFA algorithm}
    (Eqn. \eqref{eqn:MFA}); \\ 
  - Calculate $\tilde{\bm{W}}$ with $\bm{P}$, $\bm{\Lambda}$, and $\hat{\bm{H}}$
  (Eqn. \eqref{eqn:W2H}).
\end{algorithm}

\begin{figure*}
  \centering
  \includegraphics[width=0.98\linewidth]{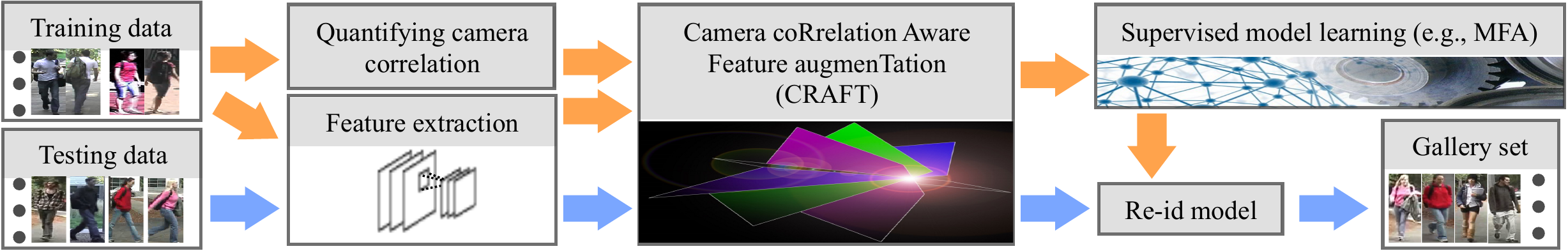}
  \vskip -.4cm
  \caption{\footnotesize
    Pipeline of the proposed person re-id approach.
    Training state is indicated with orange arrows,
    and testing stage with blue arrows.
  }
  \vspace{-.5cm}
  \label{fig:pipeline}
\end{figure*}

\subsection{CRAFT Instantialization}
In our CRAFT framework,
{we instantilize a concrete person re-id method using}
the Marginal Fisher Analysis (MFA) \cite{MFA}, 
due to
its several important advantages
over the canonical Linear Discriminant Analysis (LDA) model \cite{martinez2001pca}:
(1) no {strict} assumption on data distribution
and thus more general for discriminative learning,
(2) a much larger number of available projection directions, and
(3) a better capability of characterizing inter-class separability.
{We call this method instance as ``CRAFT-MFA''}.
Algorithm \ref{Alg:CRAFT} presents an overview of learning the CRAFT-MFA model.

Specifically, we consider each person identity as an individual class.
That is, all images of the same person form the whole same-class samples, regardless of being captured by either camera.
Formally, given the training data \cycc{$\bm{X} = [\bm{X}^a, \bm{X}^b] = [\bm{x}_1, \dots, \bm{x}_{n}]$
 with $n = n_a + n_b$}, 
we first transform them to \cycc{$\ddot{\bm{X}}_{\text{craft}} = [ \ddot{\bm{x}}_1,\cdots,\ddot{\bm{x}}_{n}]$} 
(Lines 1-9 in Algorithm \ref{Alg:CRAFT}).
$\hat{\bm{H}}$ can then be obtained
by solving the MFA optimization problem (Line \cycc{11} in Alg. \ref{Alg:CRAFT}): 
\begin{equation}
\label{eqn:MFA}
\begin{array} {ll}
\min \limits_{\bm{H}}
\sum \limits_{ i \neq j} \bm{A}^c_{ij} || \bm{H}^\top (\ddot{\bm{x}}_i - \ddot{\bm{x}}_j) ||_2^2 + \lambda \text{tr}(\bm{H}^\top \bm{H}) \\
\text{s.t.} ~~~ \sum \limits_{ i \neq j}\bm{A}^p_{ij}|| \bm{H}^\top (\ddot{\bm{x}}_i - \ddot{\bm{x}}_j) ||_2^2 = 1,
%
%
\end{array}
\end{equation}
where the element $\bm{A}^c_{ij}$ of the intrinsic graph $\bm{A}^c$ is:
\begin{equation}\label{eq:MFA:Wc}
  \bm{A}^c_{ij} = \left\{
  \begin{array}{ll}
  1 & \text{if} ~ i \in N^{+}_{k_1}(j) ~ \text{or} ~ j \in N^{+}_{k_1}(i) \\
  0 & \text{otherwise}
  \end{array} \right. \nonumber ,
\end{equation}
with $N^{+}_{k_1}(i)$ denoting the set of $k_1$ nearest neighbor indices
of sample $\bm{x}_i$ in the same class.
And the elements $\bm{A}^p_{ij}$ of the penalty graph $\bm{A}^p$ are defined as:
\begin{equation}\label{eq:MFA:Wp}
  \bm{A}^p_{ij} = \left\{
  \begin{array}{ll}
  1 & \text{if} ~ (i,j) \in P_{k_2}({y_i}) ~ \text{or} ~ (i,j) \in P_{k_2}({y_j}) \\
  0 & \text{otherwise}
  \end{array} \right. \nonumber ,
\end{equation}
{where $y_i$ and $y_j$ refer to the class/identity label of the $i^\text{th}$ and $j^\text{th}$ sample, respectively}, \cycb{$P_{k_2}(y_i)$ indicates the set of data pairs that are $k_2$ nearest pairs among 
$\{(i,j)| {y_i} \neq y_j\}$.}
%
Finally, we compute the optimal $\hat{\bm{W}}$ with Eqn. \eqref{eqn:W2H}
(Line \cycc{12} in Algorithm \ref{Alg:CRAFT}).
Note that this process generalizes to other view-generic discriminative learning algorithms \cite{martinez2001pca,LFDA,KISSME,XQDA}
(see evaluations in Table \ref{tab:eval_generality}).

\subsection{Kernel Extension}
\label{sec:kernel}

The objective function Eqn. \eqref{general_obj_3} assumes linear projection.
However, given complex variations in viewing condition across cameras,
the optimal subspace for person re-id may not be obtainable by linear models.
Thus, we further kernelize our feature augmentation (Eqn. \eqref{eq:general_transformation})
by projecting the original feature data into a reproducing kernel Hilbert space
$\mathcal{H}$ with an implicit function $\phi(\cdot)$.
The inner-product of two data points in $\mathcal{H}$
can be computed by a kernel function 
$k(\bm{x}_i, \bm{x}_j) = \left \langle \phi(\bm{x}_i), \phi(\bm{x}_j) \right \rangle$.
In our evaluation, we utilized the nonlinear Bhattacharyya kernel function
due to (1) its invariance against any non-singular linear augmentation and translation and
(2) its additional consideration of data distribution variance and thus more reliable \cite{choi2003feature}.
We denote $\bm{k}(\bm{x})$ as the kernel similarity vector of \zhuxt{a} sample \wsff{$\bm{x}$}, obtained by:
\begin{equation}\label{eq:kernel}
\begin{array}{ll}
  \bm{k}(\bm{x})
  &= [k(\bm{x}_1,\bm{x}), k(\bm{x}_2,\bm{x}), \cdots, k(\bm{x}_n,\bm{x})]^\top ,
\end{array}
\end{equation}
where \cycb{$\bm{x}_1, \bm{x}_2,\cdots,\bm{x}_n$ are all samples from all views}, \wsff{and} {$n=n_a+n_b$} is the number of training samples. Then the mapping function can be expressed as:
\begin{equation}\label{eq:kernel_map}
  f_{\text{ker}}(\bm{x}) = \bm{U}^\top \phi(\bm{X})^\top \phi(\bm{x}) = \bm{U}^\top \bm{k}(\bm{x}) ,
\end{equation}
where
$\bm{U} \in \mathbb{R}^{n}$ represents the parameter matrix to be learned.
The superscript for camera id is omitted for simplicity.
Conceptually, Eqn. \eqref{eq:kernel_map} is similar to the linear case
Eqn. \eqref{eqn:prediction} 
if we consider {$\bm{k}(\bm{x})$ as a feature representation of $\bm{x}$}.
Hence, by following 
Eqn. \eqref{eq:general_transformation},
the kernelized version of our feature augmentation can be represented as:
\begin{equation} \label{eq:kernel_transformation}
\begin{array}{ll}
\tilde{\bm{k}}^a_{\text{craft}}(\bm{x}) =
\begin{bmatrix}
\bm{R} \\[0.3em]
\bm{M}
\end{bmatrix}
{\bm{k}^a(\bm{x})}, \;\;\;\;\;
\tilde{\bm{k}}^b_{\text{craft}}(\bm{x}) =
\begin{bmatrix}
\bm{M} \\[0.3em]
\bm{R}
\end{bmatrix}
{\bm{k}^b(\bm{x})},
\end{array}
\end{equation}
where  $\bm{R} \in \mathbb{R}^{n \times n}$ and $\bm{M} \in \mathbb{R}^{n \times n}$ are the same as in Eqn. \eqref{eq:def:myPQ} but with a different dimension,
{$\bm{k}^a(\bm{x})$ and $\bm{k}^b(\bm{x})$} denote the sample kernel similarity vectors for 
camera $a$ and $b$, respectively.
Analogously, both view discrepancy regularization (Eqn. \eqref{eqn:reg}) 
and model optimization (Eqn. \eqref{general_obj_3})
can be performed similarly as in the linear case.

\vspace{-0.4cm}
\subsection{Extension to More than Two Camera Views}
\label{sec:multi-view}
There often exist \zhu{multiple} (more than two) cameras deployed across a typical surveillance network. \wsff{Suppose there are $J$($>2$) non-overlapping camera views}.
Therefore, person re-id across the whole network is realistically critical,
but \wsf{joint quantification on person association \zhu{across multiple} camera views is} largely under-studied in the current literature
\cite{Chakraborty2015Network,ma2014person}.
Compared to camera pair based re-id above,
this is a more challenging situation due to:
(1) intrinsic difference between distinct camera pairs in viewing conditions
which makes the general cross-camera feature mapping more complex and difficult to learn;
(2) quantifying simultaneously the correlations of 
multiple camera pairs is non-trivial in both formulation and computation.
In this work, we propose to address this challenge by
jointly learning adaptive view-specific re-id models
for all cameras in a unified fashion.
%
%

Specifically, we generalize
our camera correlation aware
feature augmentation (Eqn. \eqref{eq:general_transformation})
into multiple ($\wsff{J}$) camera cases as:
%
%
\begin{equation}\label{eq:general_transformation_extend}
\begin{array} {ll}
\tilde{\bm{x}}^{\phi}_{\text{craft}} = &
[\underbrace{\bm{M}_{i,1},\bm{M}_{i,2},\cdots,\bm{M}_{i,i-1}}_{\#: \; i-1}, \bm{R}_{i},
\\ [0.5cm]
& \underbrace{\bm{M}_{i,i+1},\bm{M}_{i,i+2},\cdots,\bm{M}_{i,\cycb{J}}}_{\#: \; \cycb{J}-i}]^\top \bm{x}^\phi ,
\end{array}
\end{equation}
with
\begin{equation} \label{eq:extend_M}
  \bm{M}_{i,j} = \frac{\omega_{i,j}}{\varpi_i} \bm{I}_{d \times d},
\end{equation}
where $\omega_{i,j}$ denotes the correlation between
camera $i$ and $j$, estimated by Eqn. \eqref{eq:estimate_r}.
%
Similar to Eqn. \eqref{eq:def:myPQ}, we design 
\begin{equation} \label{eq:extend_R}
\begin{array} {ll}
\bm{R}_{i} = \frac{2 - \frac{1}{\cycb{J} - 1} \sum_{j\neq i} \omega_{i,j}}{\varpi_i}
\end{array},
\end{equation}
with
\begin{equation} \label{eq:extend_varpi}
\begin{array} {ll}
{\varpi}_i = \sqrt{(2 - \frac{1}{\cycb{J}-1}\sum \limits_{j\neq i} \omega_{i,j})^2 + \sum \limits_{j \neq i} \omega_{i,j}^2}
\end{array}.
\end{equation}
%
%
Similarly, we extend the CVD regularization 
Eqn. \eqref{eqn:model_reg} to:
\begin{equation}
  \gamma_{\text{cvd}} = \sum \limits_{i,j \in \{1,2,\dots,\cycb{J}\} \; \text{and} \; i \neq j}  ||\bm{W}\zhuxt{^i} - \bm{W}\zhuxt{^j}||^2 .
  \label{eqn:model_reg_ext}
\end{equation}
Thus, the matrix $\bm{C}$ of $\gamma$ (Eqn. \eqref{eqn:reg}) is expanded as:
\begin{equation}\label{eq:mirror_B_extend}
\bm{C} =
\left(
\begin{array}{ccccc}
\bm{I} & -\beta'\bm{I} & -\beta'\bm{I} & \cdots & -\beta'\bm{I} \\
-\beta'\bm{I} & \bm{I} & -\beta'\bm{I} & \cdots & -\beta'\bm{I} \\
\vdots & \vdots & \vdots & \vdots \\
-\beta'\bm{I} & -\beta'\bm{I} & -\beta'\bm{I} & \cdots & \bm{I} \\
\end{array}
\right),
\end{equation}
%
%
%
where $\beta'=\frac{\beta}{\cycb{J}-1}$. The following view discrepancy regularized transformation and model optimization
can be carried out in the same way as in Sec. \ref{sec:reg}.
Clearly, re-id between two cameras is a special case of the extended model when $\cycb{J} = 2$.
Therefore, 
our extended CRAFT method allows to
consider and model the intrinsic correlation between camera views in
the entire network.

%
%
%

\subsection{Person Re-identification by CRAFT}
\label{sec:ReID}

Once a discriminative model ($\bm{W}$ or $\bm{U}$)
is learned from the training data using the proposed method,
we can deploy it for person re-id.
Particularly, first,
we perform feature augmentation to transform
all original features $\{\bm{x}\}$ to
the CRAFT space $\{\tilde{\bm{x}}_{\text{craft}}\}$ (Eqn. \eqref{eq:general_transformation_extend}).
Second, we match a given probe person $\tilde{\bm{x}}^p_{\text{craft}}$ from one camera
against a set of gallery people $\{\tilde{\bm{x}}_{\text{craft},i}^g\}$ from another camera
by computing the Euclidean distance
$\{\text{dist}(\tilde{\bm{x}}^p_{\text{craft}}, \tilde{\bm{x}}_{\text{craft},i}^g)\}$ in the prediction space
(induced by Eqn. \eqref{eqn:prediction} or \eqref{eq:kernel_map}).
Finally, the gallery people are sorted in {ascendant} order of their assigned distances
to generate the ranking list.
Ideally, the true match(es) can be found among a few top ranks.
The pipeline of our proposed re-id approach is depicted in Figure \ref{fig:pipeline}.


\vspace{-0.4cm}
\section{Experiments}
\label{sec:Exp}

\subsection{Datasets and Evaluation Settings}
\label{sec:datasets}

\begin{figure}
  \centering
  \subfigure[\footnotesize VIPeR ]{ 
    \includegraphics[height=0.12\textwidth] {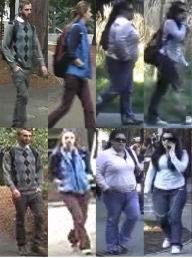}}
  \subfigure[\footnotesize CUHK01]{ 
    \includegraphics[height=0.12\textwidth] {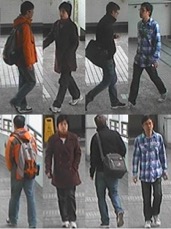}}
    \subfigure[\footnotesize CUHK03]{ 
      \includegraphics[height=0.12\textwidth] {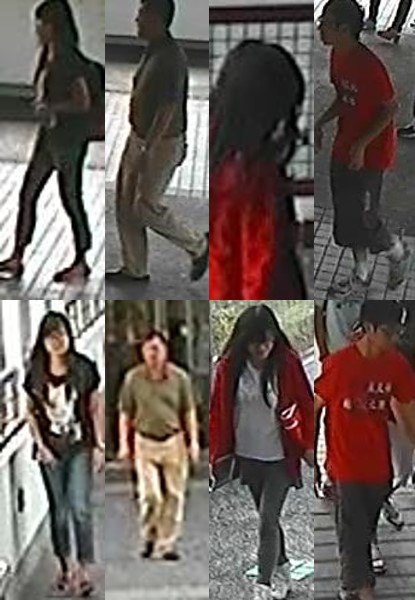}}
  \subfigure[\footnotesize GRID]{ 
    \includegraphics[height=0.12\textwidth] {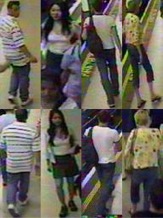}}
  \subfigure[\footnotesize Market]{ 
    \includegraphics[height=0.12\textwidth] {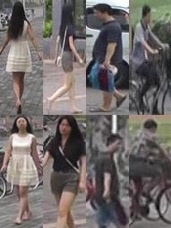}}
  \vskip -.3cm
  \caption{\footnotesize
    Example images from different person re-id datasets.
    For every dataset, two images in a column correspond to the same person.
  }
  \vspace{-.6cm}
  \label{examples:figs}
\end{figure}

\noindent \textbf{Datasets. }
We extensively evaluated the proposed approach on five person re-id benchmarks:
VIPeR \cite{viper}, CUHK01 \cite{transferREID}, {CUHK03} \cite{Deepreid},
QMUL GRID \cite{loy2009multi}, and
Market-1501 \cite{market}.
All datasets are very challenging due to unknown large cross-camera divergence in viewing conditions,
e.g., illumination, viewpoint, occlusion and background clutter (Figure \ref{examples:figs}).
%
The {\em VIPeR} dataset consists of 1264 images from 632 persons observed from two surveillance cameras
with various viewpoints and background settings.
As these images are of low spatial resolution, it is very difficult to extract
reliable appearance features 
(Figure \ref{examples:figs}(a)).
%
The {\em CUHK01} dataset consists of 971 people observed from two outdoor cameras.
Each person has two samples per camera view.
Compared with VIPeR, this dataset has higher spatial resolution and thus more
appearance information are preserved in images
(Figure \ref{examples:figs}(b)).
\xpami{The {\em CUHK03} dataset consists of 13164 images from 1360 people collected 
  from six non-overlapping cameras. 
  In evaluation, we used the automatically detected person images 
  which represent a more realistic yet challenging
  deployment scenario, e.g., due to more severe misalignment caused by imperfect detection
  (Figure \ref{examples:figs}(c)).} 
%
The {\em QMUL GRID} dataset consists of 250 people image pairs
from eight different camera views in a busy underground station.
Unlike all the three datasets above, there are 775 extra identities or imposters.
All images of this dataset are featured with
lower spatial resolution and more drastic illumination variation
(Figure \ref{examples:figs}(d)).
%
The {\em Market-1501} dataset contains person images collected
in front of a campus supermarket at Tsinghua University.
A total of six cameras were used, with five high-resolution and one low-resolution.
This dataset consists of 32,668 annotated bounding boxes of 1501 people
(Figure \ref{examples:figs}(e)).

\vspace{0.1cm}
\noindent \textbf{Evaluation protocol. }
For all datasets, we followed the standard evaluation settings
for performing a fair comparison with existing methods as below:
%
{\bf (I)} On the VIPeR, CUHK01 and QMUL GRID benchmarks,
we split randomly the whole people into two halves:
one for training and one for testing.
The cumulative matching characteristic (CMC) curve was utilized to measure the performance of {the} compared methods.
As CUHK01 is a multi-shot (e.g., multiple images per person per camera view) dataset,
we computed the final matching distance between two people by averaging corresponding cross-view image pairs.
We repeated the experiments for 10 trials and reported the average results.
\xpami{{\bf (II)} On CUHK03,
  we followed the standard protocol \cite{Deepreid} --
  repeating 20 times of random 1260/100 people splits for model training/test and 
  comparing the averaged matching results.
}
{\bf (III)} On Market-1501,
we utilized the standard training (750) and testing (751) people split provided by the authors of \cite{market}.
{Apart from CMC, we also used other two performance metrics:
  (1) Rank-1 accuracy, and (2) mean Average Precision (mAP),}
{i.e.} first computing the area under the Precision-Recall curve for each probe,
then calculating the mean of Average Precision over all probes.




 \vspace{-0.3cm}
\subsection{Evaluating Our Proposed Re-Id Approach}
\label{sec:eval_proposed}

We evaluated and analyzed the proposed person re-id approach in these aspects:
(1) Effect of our CRAFT framework 
(Eqns. \eqref{eq:general_transformation} and \eqref{eq:def:myPQ});
{(2) Comparison between CRAFT and domain adaptation;}
(3) Effect of our CVD regularization $\gamma_\text{cvd}$ (Eqn. \eqref{eqn:model_reg});
(4) Generality of {CRAFT instantialization};
(5) Effect of our HIPHOP feature;
(6) Complementary of HIPHOP on existing \wsf{popular} features.

\begin{table*}[t]
  \scriptsize
  \renewcommand{\arraystretch}{0.9}
  \setlength{\tabcolsep}{.17cm}
  \caption{
  	\footnotesize
    Comparing top matching ranks (\%) on VIPeR and CUHK01.
    The $1^\text{st}$ and $2^\text{nd}$ best results are indicated in
    \dr{red} and \db{blue} color respectively.
    }
  \label{tab:eval_feat_transformation}
  \centering
  \vskip -.4cm
  \begin{tabular}
    {c|c|c|c|c|c|c|c|c|c|c|c|c|c|c|c|c|c|c|c|c}
  \hline
  Dataset & \multicolumn{4}{c}{VIPeR \cite{viper}}
  & \multicolumn{4}{|c}{CUHK01 \cite{transferREID}} 
  & \multicolumn{4}{|c}{\xpami{CUHK03} \cite{Deepreid}} 
  & \multicolumn{4}{|c}{Market-1501 \cite{market}}
  & \multicolumn{4}{|c}{QMUL GRID \cite{loy2009multi}} \\
  \hline
  Rank \zhu{(\%)} & 1 & 5 & 10 & 20 
                  & 1 & 5 & 10 & 20 
                  & 1 & 5 & 10 & 20 
                  & 1 & 5 & 10 & 20
                  & 1 & 5 & 10 & 20\\
  \hline \hline
  OriFeat
  & 43.3 & 72.7 & 84.1 & 93.4
  & 64.3 & 85.1 & \db{90.6} & 94.6  
  & {63.4} & {88.0} & {93.0} & {96.1}
  & \db{65.4} & \db{84.0} & \db{89.3} & \db{93.1} 
  & \db{21.0} & {42.9} & {53.0} & {62.7} \\
  + Kernelization
  & 47.0 & 75.4 & 86.8 & 94.4
  & 69.5 & 89.3 & 93.5 & \db{96.5} 
  & {78.6} & {94.9} & {96.8} & {98.4}
  & \db{66.0} & \db{84.4} & \db{89.3} & \db{93.2}
  & \db{19.0} & 42.2 & 51.9 & 61.4 \\
  \hline
  ZeroPad\cite{muquet2002cyclic}
  & 37.5 & 69.9 & 82.8 & 92.1
  & \db{66.4} & \db{85.8} & 90.5 & \db{94.7}
  & 76.0 & 91.9 & 94.8 & 95.3 
  & 38.2 & 62.5 & 71.7 & 80.0 
  & 7.2 & 26.0 & 40.3 & 55.8 \\
  + Kernelization
  & 40.0 & 72.8 & 85.0 & 93.4
  & \db{71.8} & \db{89.6} & \db{93.8} & 96.5 
  & 80.0 & 92.7 & 94.4 & 95.3
  & 49.5 & 72.4 & 80.0 & 85.8
  & 6.1 & 21.8 & 36.3 & 51.4  \\ 
  \hline
  BaseFeatAug\cite{DAzeroPadding}
  & \db{45.5} & \db{76.1} & \db{87.4} & \db{95.1}
  & 59.0 & 81.1 & 87.0 & 92.4 
  & \db{78.3} & \db{94.6} & \db{97.3} & \dr{98.9} 
  & 65.3 & 83.6 & 88.8 & 92.6 
  & 20.1 & \db{47.6} & \db{58.8} & \db{70.0}\\
  + Kernelization
  & \db{47.3} & \db{77.8} & \db{89.0} & \db{95.2}
  & 63.0 & 83.5 & 89.0 & 93.6 
  & \db{83.4} & \db{97.0} & \db{98.1} & \dr{99.1}
  & 65.3 & 83.6 & 88.8 & 92.6 
  & 20.1 & \db{47.6} & \db{58.8} & \db{70.0} \\
  \hline
  {\bf CRAFT}
  & \dr{47.8} & \dr{77.1} & \dr{87.8} & \dr{95.1}
  & \dr{70.0} & \dr{87.4} & \dr{92.0} & \dr{95.5} 
  & \dr{78.5} & \dr{94.7} & \dr{97.5} & \dr{98.9}
  & \dr{67.9} & \dr{85.1} & \dr{90.0} & \dr{93.4} 
  & \dr{25.4} & \dr{50.2} & \dr{61.8} & \dr{74.2} \\
  + Kernelization
  & \dr{50.3} & \dr{80.0} & \dr{89.6} & \dr{95.5}
  & \dr{74.5} & \dr{91.2} & \dr{94.8} & \dr{97.1}
  & \dr{84.3} & \dr{97.1} & \dr{98.3} & \dr{99.1}
  & \dr{68.7} & \dr{87.1} & \dr{90.8} & \dr{94.0}
  & \dr{22.4} & \dr{49.9} & \dr{61.8} & \dr{71.7}\\
  \hline
  \end{tabular}
  \vspace{-.45cm}  
\end{table*}

\vspace{0.1cm}
\noindent {\bf Effect of our CRAFT framework. }
We evaluated the proposed CRAFT method
by comparing with
(a) {\em baseline feature augmentation} (BaseFeatAug) \cite{DAzeroPadding},
(b) {\em zero padding} (ZeroPad) \cite{muquet2002cyclic}, and
(c) {\em original features} (OriFeat).
Their kernelized \wsf{versions}
using the Bhattacharyya kernel function
\wsf{were} also evaluated.
\wsf{For fair comparison,} \zhu{our CRAFT-MFA method is utilized
  for re-id model learning}
in all compared methods.

Table \ref{tab:eval_feat_transformation} shows the re-id results.
It is evident that our CRAFT approach outperformed consistently each baseline method
on all the four re-id datasets, either using kernel or not.
For example, CRAFT surpasses OriFeat, ZeroPad and BaseFeatAug
by $4.5\% / 5.7\% /15.1\% / 2.5\% / 4.4\%$,
$10.3\% / 3.6\% / 2.5\% / 29.7\% / 18.2\%$, $2.3\% / 11.0\% / 0.2\% / 2.6\%$ $ / 5.3\%$
\wsff{at} rank-1
on VIPeR/CUHK01/CUHK03/Market-1501/QMUL GRID, respectively.
Similar improvements were observed in the kernelized case.
%
It is also found that, without feature augmentation,
OriFeat produced reasonably good person matching accuracies,
whilst both ZeroPad and BaseFeatAug may deteriorate the re-id performance.
This plausible reasons are:
(1) The small training data {size} may lead to some degree of model overfitting
given two or three times as many parameters as OriFeat; and
(2) Ignoring camera correlation can result in sub-optimal discrimination models.
%
The re-id performance can be further boosted using the kernel trick
\wsf{in} most cases except QMUL GRID.
This exception may be due to the per-camera image imbalance problem on this dataset,
e.g., $25$ images from the $6^\text{th}$ camera {\em versus} $513$ images from the $5^\text{th}$ camera.
In the \wsf{following}, we used the kernel version of all methods unless otherwise stated.

\begin{table*}[t]
  \scriptsize
  \renewcommand{\arraystretch}{0.9}
  \setlength{\tabcolsep}{.17cm}
  \caption{
  	\footnotesize
    Evaluating the effect of our CVD regularization.
    }
  \label{tab:eval_regl}
  \centering
  \vskip -.4cm
  \begin{tabular}
    {c|c|c|c|c|c|c|c|c|c|c|c|c|c|c|c|c|c|c|c|c}
    \hline
    Dataset & \multicolumn{4}{c}{VIPeR \cite{viper}}
    & \multicolumn{4}{|c}{CUHK01 \cite{transferREID}} 
    & \multicolumn{4}{|c}{\xpami{CUHK03} \cite{Deepreid}} 
    & \multicolumn{4}{|c}{Market-1501 \cite{market}}
    & \multicolumn{4}{|c}{QMUL GRID \cite{loy2009multi}} \\
    \hline
    Rank \wsf{(\%)} & 1 & 5 & 10 & 20 
                    & 1 & 5 & 10 & 20 
                    & 1 & 5 & 10 & 20 
                    & 1 & 5 & 10 & 20 
                    & 1 & 5 & 10 & 20 \\
    \hline \hline
    CRAFT(no $\gamma_\text{cvd}$)
    & 46.3 & 77.9 & 88.1 & 95.4 
    & 73.8 & 90.6 & 94.2 & 96.9 
    & 83.9 & 97.0 & 98.2 & {\bf 99.1}
    & 66.6 & 85.9 & 90.7 & 93.7 
    & 15.8 & 45.0 & 57.7 & 60.0\\
    \hline
    CRAFT 
    & {\bf 50.3} & {\bf 80.0} & {\bf 89.6} & {\bf 95.5}
    & {\bf 74.5} & {\bf 91.2} & {\bf 94.8} & {\bf 97.1}  
    & {\bf 84.3} & {\bf 97.1} & {\bf 98.3} & {\bf 99.1} 
    & {\bf 68.7} & {\bf 87.1} & {\bf 90.8} & {\bf 94.0}
    & {\bf 22.4} & {\bf 49.9} & {\bf 61.8} & {\bf 71.7}   \\
    \hline
  \end{tabular}
  \vspace{-.45cm}  
\end{table*}

\vspace{0.1cm}
\noindent {\bf {Comparison between CRAFT and domain adaptation.} }
{We {compared} our CRAFT with two \wsf{representative} domain adaptation models:
  TCA \cite{TCA} and TFLDA \cite{TFLDA}.
  It is evident from Table \ref{tab:eval_DATF} that the proposed CRAFT
  always surpasses TCA and TFLDA \wsf{with a clear margin} in the re-id performance
  on all datasets.
  This is because: (1) TCA is not discriminant and thus yielded much poor accuracies; and
  (2) both TCA and TFLDA assume that the target domain shares the same class labels as the source domain, which however is not valid in re-id applications.
   This suggests the advantage and superiority of our cross-view adaptive modelling over
   conventional domain adaptation \wsf{applied to} person re-id.
   }

\begin{table*}[t]
  \scriptsize
  \renewcommand{\arraystretch}{0.9}
  \setlength{\tabcolsep}{.18cm}
  \caption{
  	\footnotesize
    Comparison between CRAFT and domain adaptation. 
  }
  \label{tab:eval_DATF}
  \centering
  \vskip -.4cm
  \begin{tabular}
    {c|c|c|c|c|c|c|c|c|c|c|c|c|c|c|c|c|c|c|c|c}
    \hline
    Dataset & \multicolumn{4}{c}{VIPeR \cite{viper}}
    & \multicolumn{4}{|c}{CUHK01 \cite{transferREID}} 
    & \multicolumn{4}{|c}{\xpami{CUHK03} \cite{Deepreid}}
    & \multicolumn{4}{|c}{Market-1501 \cite{market}}
    & \multicolumn{4}{|c}{QMUL GRID \cite{loy2009multi}}\\
    \hline
    Rank \wsf{(\%)} & 1 & 5 & 10 & 20 
                    & 1 & 5 & 10 & 20 
                    & 1 & 5 & 10 & 20 
                    & 1 & 5 & 10 & 20 
                    & 1 & 5 & 10 & 20 \\
    \hline \hline
    TCA \cite{TCA}
    & 11.1 & 23.4 & 31.0 & 38.5 
    & 7.0 & 16.4 & 22.2 & 30.1 
    & 5.5 & 16.2 & 26.4 & 42.8
    & 8.9 & 18.7 & 24.1 & 30.1 
    & 9.8 & 22.2 & 29.8 & 38.3\\
    \hline
    TFLDA \cite{TFLDA}
    & 46.4 & 75.8 & 86.7 & 93.9 
    & 69.6 & 88.7 & 92.8 & 96.2 
    & 76.7 & 94.4 & 96.5 & 98.0
    & 62.5 & 81.3 & 87.0 & 91.6 
    & 19.5 & 42.5 & 51.6 & 61.8\\
    \hline
    {\bf CRAFT}
    & {\bf 50.3} & {\bf 80.0} & {\bf 89.6} & {\bf 95.5}
    & {\bf 74.5} & {\bf 91.2} & {\bf 94.8} & {\bf 97.1}  
    & {\bf 84.3} & {\bf 97.1} & {\bf 98.3} & {\bf 99.1} 
    & {\bf 68.7} & {\bf 87.1} & {\bf 90.8} & {\bf 94.0}
    & {\bf 22.4} & {\bf 49.9} & {\bf 61.8} & {\bf 71.7}   \\
    \hline
  \end{tabular}
  \vspace{-.45cm}  
\end{table*}

\vspace{0.1cm}
\noindent {\bf Effect of our CVD regularization.}
For evaluating the exact impact of the proposed regularization $\gamma_\text{cvd}$ (Eqns. \eqref{eqn:model_reg} and \eqref{eqn:model_reg_ext}) on model generalization,
we compared the re-id performance of our full model with
a stripped down variant
without $\gamma_\text{cvd}$ during model optimization,
called ``CRAFT(no $\gamma_\text{cvd}$)''.
The results in Table \ref{tab:eval_regl} demonstrate
the \wsf{usefulness} of incorporating $\gamma_\text{cvd}$
for re-id.
\wsf{This justifies the effectiveness of our CVD regularization in controlling the correlation degree between view-specific sub-models.}

\begin{table*}[t] \scriptsize
    \renewcommand{\arraystretch}{0.9}
  \setlength{\tabcolsep}{.25cm}
  \caption{
  	\footnotesize 
    Evaluating the generality of {CRAFT instantialization.}
    In each row, the $1^\text{st}/2^\text{nd}$ best results (\%) for each rank are indicated
    in \dr{red} / \db{blue} color.
  }
  \label{tab:eval_generality}
  \centering
  \vskip -0.4cm
  \begin{tabular}{c|c|c|c|c|c|c|c|c|c|c|c|c|c|c|c|c|c}
    \hline
    & Method
    & \multicolumn{4}{c}{\bf CRAFT}
    & \multicolumn{4}{|c}{OriFeat}
    & \multicolumn{4}{|c}{ZeroPad\cite{muquet2002cyclic}}
    & \multicolumn{4}{|c}{BaseFeatAug\cite{DAzeroPadding}}
    \\
    \hline
    & Rank \zhu{(\%)} & 1 & 5 & 10 & 20 & 1 & 5 & 10 & 20 & 1 & 5 & 10 & 20 & 1 & 5 & 10 & 20 \\
    \hline \hline
    
    \multirow{5}{*}{\begin{sideways}  VIPeR\end{sideways}} 
    & MFA \cite{MFA}
    & \dr{50.3} & \dr{80.0} & \dr{89.6} & \dr{95.5} & 47.0 & 75.4 & 86.8 & 94.4 & 40.0 & 72.8 & 85.0 & 93.4 & \db{47.3} & \db{77.8} & \db{89.0} & \db{95.2} \\ \cline{2-18}
    & LDA \cite{martinez2001pca}
    & \dr{49.3} & \dr{79.1} & \dr{89.2} & \dr{95.3} & 46.4 & 75.0 & 86.5 & 94.1 & 40.0 & 73.8 & 84.7 & 93.3 & \db{47.4} & \db{78.0} & \db{88.9} & \db{95.3} \\ \cline{2-18}
    & KISSME \cite{KISSME}
    & \dr{50.1} & \dr{79.1} & \dr{88.9} & \dr{95.0} & 46.3 & 75.1 & 86.5 & 94.0 & 38.7 & 71.1 & 83.8 & 92.3 & \db{48.4} & \db{77.6} & \db{88.7} & \db{94.9} \\ \cline{2-18}
    & LFDA \cite{LFDA}
    & \dr{49.3} & \dr{79.1} & \dr{89.2} & \dr{95.3} & 46.4 & 75.0 & 86.5 & 94.1 & 40.0 & 73.8 & 84.7 & 93.3 & \db{47.4} & \db{77.7} & \db{88.8} & \db{95.2} \\ \cline{2-18}
    & XQDA \cite{XQDA}
    & \dr{50.3} & \dr{79.1} & \dr{88.9} & \dr{95.0} & 46.3 & 75.1 & 86.5 & 94.0 & 38.7 & 71.1 & 83.8 & 92.3 & \db{47.1} & \db{76.9} & \db{88.9} & \db{94.7} \\
    \hline \hline
    \multirow{5}{*}{\begin{sideways}CUHK01\end{sideways}} 
    & MFA \cite{MFA}
    & \dr{74.5} & \dr{91.2} & \dr{94.8} & \dr{97.1} & 69.5 & 89.3 & 93.5 & \db{96.5} & \db{71.8} & \db{89.6} & \db{93.8} & 96.5 & 63.0 & 83.5 & 89.0 & 93.6 \\ \cline{2-18}
    & LDA \cite{martinez2001pca}
    & \dr{73.8} & \dr{90.3} & \dr{93.9} & \dr{96.6} & 69.4 & 88.3 & 92.8 & \db{96.1} & \db{71.3} & \db{88.8} & \db{92.9} & 96.1 & 63.0 & 83.5 & 88.9 & 93.4 \\ \cline{2-18}
    & KISSME \cite{KISSME}
    & \dr{73.0} & \dr{89.6} & \dr{93.6} & \dr{96.3} & \db{69.2} & \db{87.6} & \db{92.5} & \db{95.8} & 67.3 & 85.9 & 90.8 & 94.7 & 63.6 & 83.0 & 88.0 & 92.9 \\ \cline{2-18}
    & LFDA \cite{LFDA}
    & \dr{73.8} & \dr{90.3} & \dr{93.9} & \dr{96.6} & 69.4 & 88.3 & 92.8 & \db{96.1} & \db{71.3} & \db{88.8} & \db{92.9} & 96.1 & 62.6 & 83.1 & 88.5 & 93.3 \\ \cline{2-18}
    & XQDA \cite{XQDA}
    & \dr{73.0} & \dr{89.5} & \dr{93.6} & \dr{96.3} & 69.1 & \db{87.6} & \db{92.6} & \db{95.8} & \db{69.9} & 87.3 & 92.1 & 95.5 & 61.3 & 82.2 & 87.8 & 93.1 \\
        \hline
        \hline
     \multirow{5}{*}{\begin{sideways}\xpami{CUHK03}\end{sideways}}
    & MFA \cite{MFA}
    & \dr{84.3} & \dr{97.0} & \dr{98.3} & \dr{99.1} 
    & 78.6 & 94.9 & 96.8 & 98.3
    & 80.0 & 92.7 & 94.4 & 95.3
    & \db{83.4} & \dr{97.0} & \db{98.1} & \dr{99.1} \\
    \cline{2-18}
    & LDA \cite{martinez2001pca}
    & \dr{80.2} & \dr{96.6} & \dr{98.2} & \dr{99.0} 
    & 76.6 & 94.6 & 96.6 & 98.0
    & 79.5 & 94.7 & 96.4 & 98.3
    & \db{80.0} & \db{96.2} & \db{97.3} & \db{99.0} \\
    \cline{2-18}
    & KISSME \cite{KISSME}
    & \dr{76.2} & \dr{93.7} & \dr{96.9} & \dr{98.5}
    & 64.7 & 88.7 & 93.4 & 96.2
    & \db{73.6} & \db{92.5} & \db{95.9} & \db{98.0}
    & 72.4 & 91.6 & 95.3 & 97.8 \\
    \cline{2-18}
    & LFDA \cite{LFDA}
    & \dr{80.9} & \dr{96.4} & \dr{98.3} & \dr{99.0}
    & 76.3 & 93.5 & 96.3 & 97.8
    & 79.7 & 95.6 & 97.4 & 98.0
    & \db{80.4} & \db{96.3} & \db{98.2} & \db{98.9} \\
    \cline{2-18}
    & XQDA \cite{XQDA}
    & \dr{79.8} & \dr{96.0} & \dr{98.0} & \dr{99.0}
    & 78.7 & 94.3 & 97.4 & \db{98.7}
    & 78.3 & \db{95.8} & \db{97.4} & 98.2
    & \db{79.2} & 95.5 & 97.3 & 98.1 \\
    \hline
    \hline

\multirow{6}{*}{\begin{sideways}GRID\end{sideways}} 
 & MFA \cite{MFA} & \dr{22.4} & \dr{49.9} & \dr{61.8} & \dr{71.7} & \db{19.0} & 42.2 & 51.9 & 61.4 & 6.1 & 21.8 & 36.3 & 51.4 & 17.0 & \db{45.8} & \db{58.2} & \db{67.9} \\ \cline{2-18}
& LDA \cite{martinez2001pca} & \dr{22.1} & \dr{50.1} & \dr{61.6} & \dr{71.0} & \db{19.0} & 42.2 & 51.7 & 61.7 & 6.6 & 23.4 & 37.3 & 50.7 & 17.4 & \db{46.6} & \db{57.6} & \db{68.3} \\ \cline{2-18}
& KISSME \cite{KISSME} & \dr{22.4} & \dr{50.4} & \dr{61.4} & \dr{71.5} & \db{19.5} & 41.8 & 51.6 & 61.0 & 5.5 & 21.4 & 35.8 & 50.8 & 17.0 & \db{46.5} & \db{57.4} & \db{67.5} \\ \cline{2-18}
& LFDA \cite{LFDA} & \dr{22.2} & \dr{50.1} & \dr{61.5} & \dr{71.0} & \db{19.0} & 42.0 & 51.6 & 61.7 & 6.6 & 23.4 & 37.3 & 50.7 & 17.4 & \db{46.6} & \db{57.6} & \db{68.3} \\ \cline{2-18}
& XQDA \cite{XQDA} & \dr{22.3} & \dr{50.9} & \dr{61.6} & \dr{71.4} & \db{19.8} & 42.5 & 51.7 & 61.8 & 6.1 & 22.7 & 36.5 & 51.4 & 17.0 & \db{46.5} & \db{57.4} & \db{67.8} \\
\hline
\hline
\multirow{5}{*}{\begin{sideways}Market\end{sideways}} 
 & MFA \cite{MFA} & \dr{68.7} & \dr{87.1} & \dr{90.8} & \dr{94.0} & \db{66.0} & 84.4 & 89.3 & 93.2 & 49.5 & 72.4 & 80.0 & 85.8 & 65.5 & \db{84.5} & \db{90.5} & \db{93.4} \\ \cline{2-18}
& LDA \cite{martinez2001pca} & \dr{62.9} & \dr{82.2} & \dr{88.6} & \dr{92.4} & \db{62.6} & \db{81.9} & \db{87.6} & \db{92.2} & 47.9 & 72.9 & 81.9 & 87.8 & 60.8 & 81.8 & 87.3 & 91.8 \\ \cline{2-18}
& KISSME \cite{KISSME} & \dr{61.4} & \dr{82.4} & \dr{88.5} & \dr{92.6} & \db{58.7} & \db{78.7} & \db{85.3} & \db{90.1} & 51.2 & 76.3 & 84.3 & 89.5 & 51.2 & 75.7 & 83.6 & 89.7 \\ \cline{2-18}
& LFDA \cite{LFDA} & \dr{68.0} & \dr{85.5} & \dr{90.9} & \dr{94.5} & 61.7 & 79.5 & 85.7 & 90.6 & 47.7 & 72.4 & 80.9 & 87.9 & \db{65.4} & \db{83.6} & \db{89.5} & \db{93.3} \\ \cline{2-18}
& XQDA \cite{XQDA} & \dr{61.3} & \dr{81.9} & \dr{87.6} & \dr{92.1} & \db{55.5} & 76.6 & 84.1 & 89.3 & 51.0 & \db{77.1} & \db{84.3} & \db{89.4} & 44.2 & 71.1 & 81.4 & 88.1 \\

\hline
  \end{tabular}
  \vspace{-.45cm}
\end{table*}

\vspace{0.1cm}
\noindent {\bf Generality of CRAFT instantialization. } 
We evaluated the generality of CRAFT instantialization
by integrating different supervised learning models. 
Five popular state-of-the-art methods were considered:
(1) MFA \cite{MFA},
(2) LDA \cite{martinez2001pca},
(3) KISSME \cite{KISSME},
(4) LFDA \cite{LFDA},
(5) XQDA \cite{XQDA}.
Our kernel feature was adopted.


The results are given in Table \ref{tab:eval_generality}.
It is evident that our CRAFT framework is general {for} \wsf{incorporating} different existing {distance metric} learning algorithms.
Specifically, we \wsf{find} that \wsf{CRAFT achieved better re-id results than other feature augmentation competitors
with \zhu{any of these learning algorithms.}}
\wsf{The observation also suggests the consistent superiority and
large flexibility of the proposed approach over alternatives
in learning discriminative re-id models.}
We utilize CRAFT-MFA 
for the following comparisons with existing \wsf{popular} person re-id methods.


\begin{table*}[t]
  \scriptsize
  \renewcommand{\arraystretch}{0.9}
  \setlength{\tabcolsep}{.15cm}
  \caption{\footnotesize
    Evaluating the effectiveness of our HIPHOP feature.
    Best results \wsf{(\%)} by single and fused features are indicated in \db{blue} and \dr{red} color respectively. \cyc{Our CRAFT-MFA is utilized for each type of feature in this experiment.}
    }
  \label{tab:eval_feat}
  \centering
  \vskip -.4cm
  \begin{tabular}
    {c||c|c|c|c||c|c|c|c||c|c|c|c||c|c|c|c||c|c|c|c}
    \hline
    Dataset & \multicolumn{4}{c||}{VIPeR \cite{viper}}
    & \multicolumn{4}{c||}{CUHK01 \cite{transferREID}}
    & \multicolumn{4}{c||}{\xpami{CUHK03} \cite{Deepreid}}
        & \multicolumn{4}{c||}{Market-1501 \cite{market}}
        & \multicolumn{4}{c}{QMUL GRID \cite{loy2009multi}}\\
    \hline
    Rank \zhu{(\%)} & 1 & 5 & 10 & 20 
                    & 1 & 5 & 10 & 20
                    & 1 & 5 & 10 & 20
                    & 1 & 5 & 10 & 20 
                    & 1 & 5 & 10 & 20 \\
    \hline \hline
    ELF18\cite{chen2015CVDCA}
    & 38.6 & 69.2 & 82.1 & 91.4 
    & 51.8 & 75.9 & 83.1 & 88.9
    & 65.8 & 89.9 & 93.8 & 97.0 
    & 46.4 & 70.0 & 78.3 & 84.9 
    & 19.1 & 41.8 & 55.1 & 67.0\\
    ColorLBP\cite{hirzer2012relaxed}
    & 16.1 & 36.8 & 49.4 & 64.9 
    & 32.3 & 53.1 & 62.9 & 73.0
    & 29.8 & 57.3 & 70.6 & 82.9
    & 13.6 & 28.2 & 36.2 & 45.9 
    & 5.8 & 15.0 & 22.8 & 32.2\\
    WHOS \cite{ISR}
    & 37.1 & 67.7 & 79.7 & 89.1 
    & 66.0 & 85.5 & 91.0 & 95.1
    & 72.0 & 92.9 & 96.4 & 98.4
    & 52.2 & 76.0 & 83.3 & 89.6 
    & 20.7 & 47.4 & 58.8 & 69.8\\
    LOMO\cite{XQDA}
    & 42.3 & 74.7 & 86.5 & 94.2 
    & 65.4 & 85.3 & 90.5 & 94.1
    & 77.9 & 95.5 & 97.9 & 98.9
    & 47.2 & 72.9 & 81.7 & 88.2 
    & 21.4 & 43.0 & 53.1 & 65.5\\
    \hline
    {\bf HIPHOP}
    & \db{50.3} & \db{80.0} & \db{89.6} & \db{95.5} 
    & \db{74.5} & \db{91.2} & \db{94.8} & \db{97.1}
    & \db{84.3} & \db{97.1} & \db{98.3} & \db{99.1}
    & \db{68.7} & \db{87.1} & \db{90.8} & \db{94.0} 
    & \db{22.4} & \db{49.9} & \db{61.8} & \db{71.7}\\
    
    HOP only 
    & 47.6 & 75.6 & 86.6 & 93.8 
    & 68.9 & 87.7 & 92.7 & 96.3
    & 76.4 & 89.5 & 93.6 & 97.3
    & 57.9 & 78.7 & 85.6 & 90.5 
    & 18.0 & 48.2 & 59.9 & 70.2\\
    
    HIP only 
    & 47.8 & 76.4 & 86.4 & 94.3 
    & 70.0 & 88.0 & 92.2 & 95.5
    & 79.5 & 92.3 & 96.2 & 97.4
    & 67.5 & 85.0 & 89.7 & 92.9 
    & 21.6 & 46.2 & 59.4 & 71.0\\
    
    fc6 only 
    & 15.0 & 36.6 & 50.9 & 66.0 
    & 19.6 & 42.1 & 54.0 & 66.3
    & 21.2 & 43.3 & 56.1 & 67.2
    & 13.5 & 31.3 & 42.3 & 54.1 
    & 9.6 & 21.1 & 29.3 & 40.2\\
    
    fc7 only 
    & 10.0 & 27.3 & 39.3 & 55.1 
    & 11.7 & 27.3 & 37.0 & 48.6
    & 12.7 & 27.9 & 36.2 & 48.4
    & 9.2 & 24.2 & 34.1 & 45.3 
    & 5.1 & 14.5 & 20.6 & 32.4\\
    
%
    
%
    

    \hline \hline
    {\bf HIPHOP}+ELF18
    & 52.5 & 81.9 & 91.3 & 96.5 
    & 75.6 & 91.6 & 95.1 & 97.1
    & 86.5 & \dr{97.4} & 98.6 & 99.4
    & 69.9 & 86.9 & 90.9 & 94.4 
    & 23.1 & 50.7 & 60.9 & 72.4\\
    {\bf HIPHOP}+ColorLBP
    & 51.0 & 80.3 & 90.2 & 95.7 
    & 74.3 & 90.8 & 94.4 & 96.9
    & 84.0 & 97.0 & 98.6 & \dr{99.5}
    & 69.3 & 86.9 & 91.4 & 94.1 
    & 22.9 & 48.4 & 59.2 & 70.0\\
    {\bf HIPHOP}+WHOS
    & 52.5 & 81.0 & 90.5 & 96.2 
    & 75.9 & 92.1 & 95.2 & 97.3
    & 85.0 & \dr{97.4} & \dr{98.7} & \dr{99.5}
    & 70.3 & 88.1 & 91.7 & 94.6 
    & 24.5 & \dr{53.3} & \dr{64.4} & \dr{75.0}\\
    {\bf HIPHOP}+LOMO
    & \dr{54.2} & \dr{82.4} & \dr{91.5} & \dr{96.9}
    & \dr{78.8} & \dr{92.6} & \dr{95.3} & \dr{97.8}
    & \dr{87.5} & \dr{97.4} & \dr{98.7} & \dr{99.5}
    & \dr{72.3} & \dr{88.2} & \dr{91.9} & \dr{95.0} 
    & \dr{26.0} & 50.6 & 62.5 & 73.3\\
    \hline
  \end{tabular}
  \vspace{-.5cm}  
\end{table*}

\begin{figure}[t]
  \centering
  \includegraphics[width=1\linewidth]{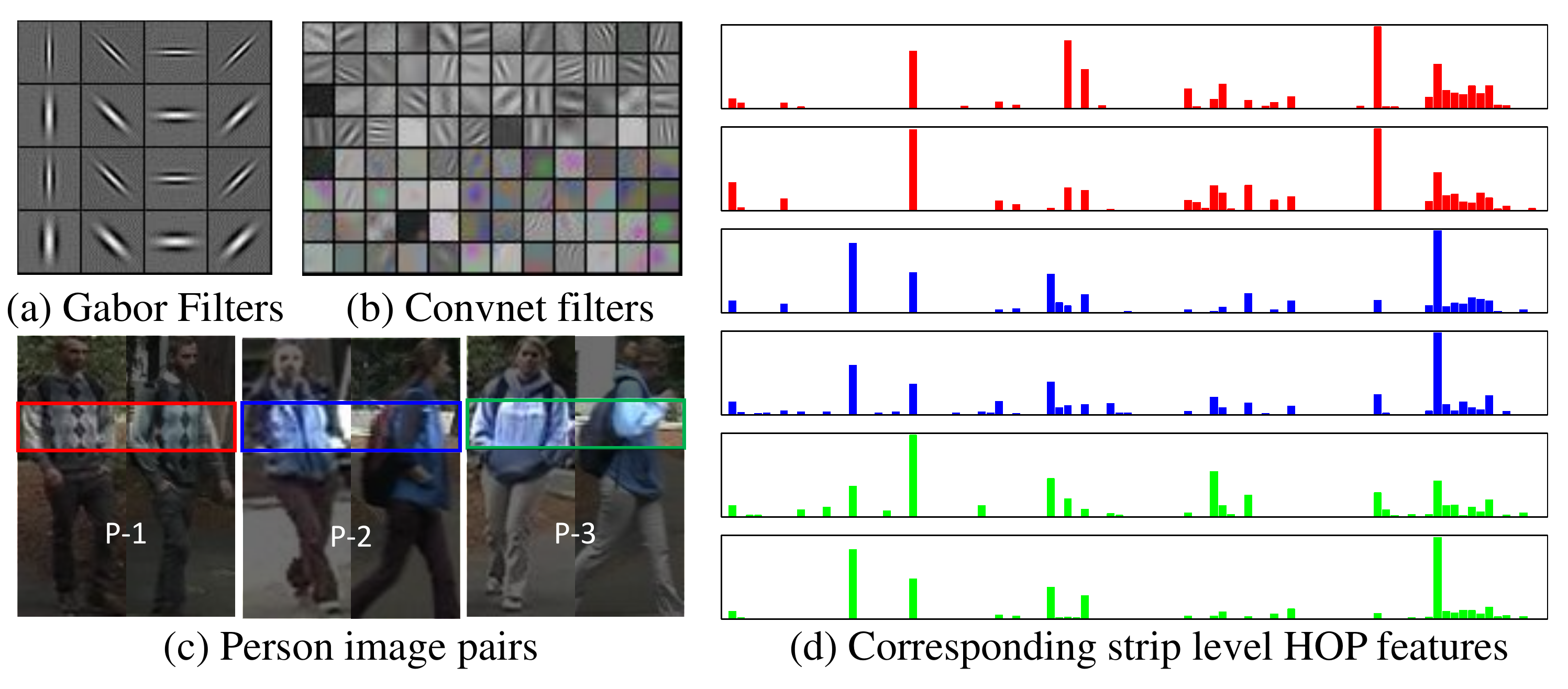}
  \vskip -0.5cm
  \caption{\footnotesize Illustration of our HIP and HOP person descriptors.
    (a) Gabor filters used in ELF18 \cite{chen2015CVDCA}.
    (b) Convolutional filters from the $1^\text{st}$ AlexNet layer.
    (c) Horizontal stripes of $6$ images from $3$ different people ({P-1, P-2 and P-3}).
    (d) HOP histograms extracted from the corresponding image strips
    (\zhu{i.e., indicated with a rectangular of the same color as histogram bars})
    in (c).
    Partial HOP descriptors are shown for clear visualization.
    }
  \vspace{-0.55cm}
  \label{fig:illu_feat}
\end{figure}

\vspace{0.1cm}
\noindent {\bf Effect of our HIPHOP feature.}
We compared the proposed HIPHOP feature with several extensively adopted re-id representations:
(1) ELF18 \cite{chen2015CVDCA},
(2) ColorLBP \cite{hirzer2012relaxed},
(3) WHOS \cite{ISR}, and
(4) LOMO \cite{XQDA}.
The Bhattacharyya kernel was utilized for all compared visual features.


The re-id results are reported in Table \ref{tab:eval_feat}.
It is evident that our HIPHOP feature is \wsf{overall} much more effective than the compared alternatives for person re-id.
For example, using HIPHOP improved the rank-1 rate
from $42.3\%$ by LOMO to $50.3\%$ on VIPeR,
from $66.0\%$ by WHOS to $74.5\%$ on CUHK01,
\xpami{from $77.9\%$ by LOMO to $84.3\%$ on CUHK03,} 
\wsf{and} from $52.2\%$ by WHOS to $68.7\%$ on Market-1501.
This suggests the great advantage and effectiveness of
more view invariant 
appearance representation 
learned from diverse \wsf{albeit} generic auxiliary data.
A few reasons are:
(1) By exploiting more complete and diverse filters (Figure \ref{fig:illu_feat}(b))
than ELF18 (Figure \ref{fig:illu_feat}(a)),
\wsf{more view change tolerant} person appearance details shall be well encoded for
discriminating between visually alike people;
(2) By selecting salient patterns,
our HOP descriptor possesses some \wsf{features more tolerant to view variety,}
which is critical to person re-id due to the potential cross-view pose \zhu{and viewing condition} variation.
This is clearly illustrated in Figure \ref{fig:illu_feat}(c-d):
(i) Cross-view images of the same person have similar HOP patterns;
(ii) Visually similar people (P-2, P-3) share more commonness than
visually distinct people (P-1, P-2) in HOP histogram.
%
%
Besides, we evaluated the discrimination power of features
from different conv layers {in} Table \ref{tab:eval_feat}.
More specifically, it is shown in Table \ref{tab:eval_feat}
that the $1^\text{st}/2^{\text{nd}}$ conv layer based features
(i.e., HIP/HOP, being low-level)
are shown more effective than those from the $6^{\text{th}}/7^{\text{th}}$ conv layer
(i.e., fc6/fc7, being abstract).
This confirms early findings \cite{decaf,visualizeCNN}
that higher layers are more task-specific,
thus less generic to distinct tasks.

Recall that the ordinal feature HOP (Eqn. \eqref{eq:hop:rank})
is computed based on top-$\kappa$ activations of the feature maps.
We further examined the effect of $\kappa$ on the feature quality on the VIPeR dataset.
For obtaining a detailed analysis, we tested the HOP feature
extracted from the $1^\text{st}/2^\text{nd}$ conv layer and both layers, separately.
Figure \ref{fig:para_hop} shows the impact of setting
different $\kappa$ values ranging from $5$ to $50$
in terms of rank-1 recognition rate.
The observations suggest clearly that
$\kappa$ is rather insensitive with a wide satisfiable range.
We set $\kappa=20$ in all other evaluations.


\begin{figure}[t]
  \centering
  \includegraphics[width=0.8\linewidth] {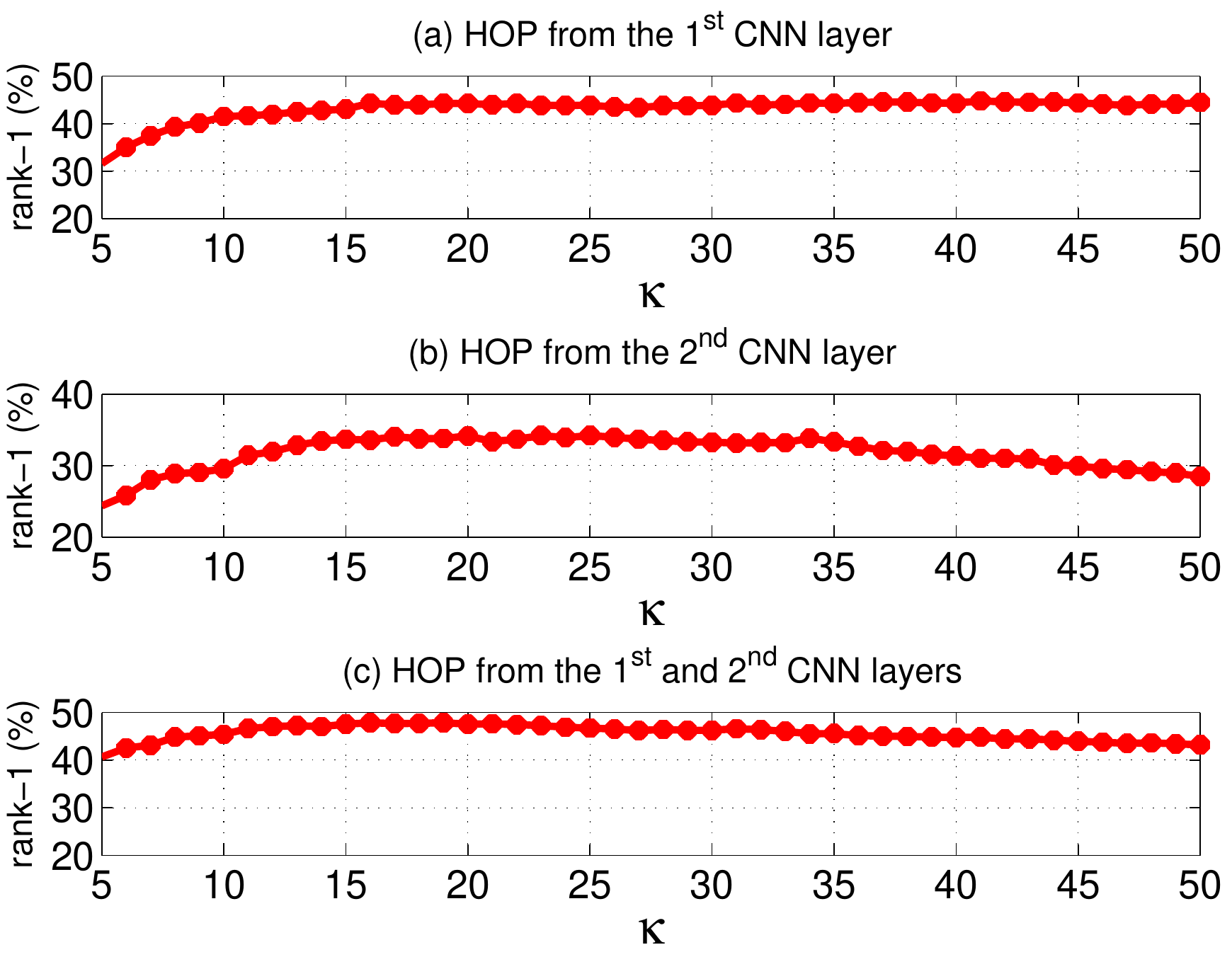}
  \vskip -.45cm
  \caption{\footnotesize
    Evaluating the effect of $\kappa$ in the HOP descriptor.
    }
  \label{fig:para_hop}
  \vspace{-0.65cm}
\end{figure}

\vspace{0.1cm}
\noindent {\bf Complementary of HIPHOP on existing re-id features. }
Considering the different design nature of our person representation
as compared to previous re-id features,
we evaluated the {complementary} effect of our HIPHOP with {ELF18, ColorLBP, WHOS and LOMO}
(see the last four rows in Table \ref{tab:eval_feat}).
It is found that:
(1) After combining our HIPHOP, all these existing features
can produce much better re-id performance.
This validates the favourable complementary role of our HIPHOP feature
for existing ones. 
(2) Interestingly, these four fusions produce rather similar and competitive re-id performance,
as opposite to the large differences in the results by each individual existing feature
(see the first four rows in Table \ref{tab:eval_feat}).
This justifies the general complementary importance of the proposed feature for
different existing re-id features.
Next, we utilize ``HIPHOP+LOMO'' as the default multi-type feature fusion
in the proposed CRAFT-MFA method due to
its slight superiority over other combinations.
This is termed as {\bf CRAFT-MFA(+LOMO)} in the remaining evaluations.

\vspace{-0.3cm}
\subsection{Comparing State-of-the-Art Re-Id Methods}
\label{sec:exp_comp_existing_methods}

We compared extensively our method CRAFT-MFA with state-of-the-art person re-id approaches.
In this evaluation, we considered two scenarios:
(1) Person re-id between two cameras;
(2) Person re-id \zhu{across multiple} ($>$$2$) cameras.
Finally, we compared the person re-id performance by
different methods using multiple feature types.
We utilized the best results reported in the corresponding papers
for fair comparison across all \wsf{compared} models.

\begin{table}[t] \scriptsize
  \renewcommand{\arraystretch}{0.9}
  \setlength{\tabcolsep}{.42cm}
  \caption{\footnotesize
    Comparing state-of-the-art methods on {\em VIPeR}\cite{viper}.
    }
  \label{tab:viper_system}
  \centering
  \vskip -0.4cm
  \begin{tabular}{c|c|c|c|c}
    \hline
    Rank (\%) & 1 & 5 & 10 & 20 \\
    \hline \hline
    RDC \cite{PRDC} & 15.7 & 38.4 & 53.9 & 70.1 \\
    \hline
    KISSME \cite{KISSME}
    & 22.0 & - & 68.0 & - \\ 
    \hline
    LFDA \cite{LFDA} & 24.2 & 52.0 & 67.1 & 82.0 \\
    \hline
    RPLM \cite{hirzer2012relaxed}
    & 27.0 & - & 69.0 & 83.0 \\
    \hline
    MtMCML \cite{ma2014person}
    & 28.8 & 59.3 & 75.8 & 88.5 \\
        \hline
    LADF \cite{LADF} & 29.3 & 61.0 & 76.0 & 88.1 \\
    \hline
    SalMatch \cite{salienceMatch}
    & 30.2 & 52.3 & 65.5 & 79.2 \\
    \hline
    MFA \cite{kernelREIDECCV14}
    & 32.2 & 66.0 & 79.7 & 90.6 \\ 
    \hline
    kLFDA \cite{kernelREIDECCV14}
    & 32.3 & 65.8 & 79.7 & 90.9 \\ 
    \hline
        Ref-reid \cite{an2015person}
        & 33.3 & - & 78.4 & 88.5 \\
        \hline
    SCNCD \cite{colorName}
    & 33.7 & 62.7 & 74.8 & 85.0 \\
    \hline
    Siamese-Net \cite{yi2014deep}
    & 34.4 & 62.2 & 75.9 & 87.2 \\
    \hline
    CIND-Net \cite{ahmed2015improved}
    & 34.8 & 63.6 & 75.6 & 84.5 \\
    \hline
    CorStruct \cite{shen2015person}
    & 34.8 & 68.7 & 82.3 & 91.8 \\
        \hline
    PolyMap \cite{chen2015similarity}
    & 36.8 & 70.4 & 83.7 & 91.7 \\
    \hline
    KCCA \cite{lisanti2014matching}
    & 37.0 & - & 85.0 & 93.0 \\
    \hline
    \xpami{DGD} \cite{Dropout_reid} & 38.6 & - & - & - \\
    \hline
    XQDA \cite{XQDA}
    & {40.0} & {68.1} & {80.5} & {91.1} \\
    \hline
    MLAPG \cite{LiaoPSD}
    & 40.7 & 69.9 & 82.3 & 92.4\\
    \hline
    RDC-Net\cite{ding2015deep}
    & 40.5 & 60.8 & 70.4 & 84.4 \\
    \hline
    DNS \cite{zhang2016learning}
    & 42.3 & 71.5 & 82.9 & 92.1 \\
    \hline
    KEPLER \cite{KEPLER}
    & 42.4 & - & 82.4 & 90.7\\
    \hline
    \xpami{LSSCDL} \cite{LSSCDL} 
    & 42.7 & - & 84.3 & 91.9 \\
    \hline
    CVDCA \cite{chen2015CVDCA}
    & 43.3 & 72.7 & 83.5 & 92.2 \\
    \hline
    Metric Ensemble \cite{ensembleReID}
    & 45.9 & 77.5 & 88.9 & 95.8  \\
    \hline
    \xpami{TCP} \cite{Cheng_TCP}
    & 47.8 & 74.7 & 84.8 & 89.2 \\
    \hline \hline
    {\bf CRAFT-MFA}
    & \bf{50.3} & \bf{80.0} & \bf{89.6} & \bf{95.5} \\
        \hline
  \end{tabular}
  \vspace{-.3cm}
\end{table}

\vspace{0.1cm}
\noindent {\bf (I) Person re-id between two cameras. }
This evaluation was carried out on
VIPeR \cite{viper} and CUHK01 \cite{transferREID} datasets,
each with a pair of camera views.
We compared our CRAFT-MFA with both metric learning
and recent deep learning based methods.

\vspace{0.1cm}
\noindent {\em Comparisons on VIPeR - }
We compared with $26$ state-of-the-art re-id methods on VIPeR.
The performance comparison is reported in Table \ref{tab:viper_system}.
It is evident that our CRAFT-MFA method surpassed all competitors
over top ranks clearly.
\xpami{For instance, the rank-1 rate is improved notably
from $47.8\%$ (by the $2^\text{nd}$ best method TCP \cite{Cheng_TCP}) to $50.3\%$.}
This shows the advantages and effectiveness of our approach in a broad context.

\begin{table}[t] 
	\scriptsize
	\centering
  \renewcommand{\arraystretch}{0.9}
  \setlength{\tabcolsep}{.43cm}
  \caption{\footnotesize
    Comparing state-of-the-art methods on {\em CUHK01}\cite{transferREID}.
    }
    \label{tab:cuhk01_system}
  \vskip -0.4cm
  \centering
  \begin{tabular}{c|c|c|c|c}
    \hline
    Rank (\%) & 1 & 5 & 10 & 20 \\
    \hline \hline
    LMNN \cite{LMNN}
    & 13.4 & 31.3 & 42.3 & 54.1 \\
    \hline
    ITML \cite{ITML}
    & 16.0 & 35.2 & 45.6 & 59.8 \\
    \hline
    eSDC \cite{eSDCocsvm}
    & 19.7 & 32.7 & 40.3 & 50.6 \\ 
    \hline
    GM \cite{transferREID}
    & 20.0 & - & 56.0 & 69.3 \\
    \hline
    SalMatch \cite{salienceMatch}
    & 28.5 & 45.9 & 55.7 & 68.0 \\ 
    \hline
        Ref-reid \cite{an2015person}
        & 31.1 & - & 68.6 & 79.2 \\
        \hline
    MLF \cite{medFilter}
    & 34.3 & 55.1 & 65.0 & 74.9 \\ 
    \hline
    CIND-Net \cite{ahmed2015improved} 
    & 47.5 & 71.6 & 80.3 & 87.5 \\ 
    \hline
    CVDCA \cite{chen2015CVDCA}
    & 47.8 & 74.2 & 83.4 & 89.9 \\
    \hline
      Metric Ensemble \cite{ensembleReID}
    & 53.4 & 76.4 & 84.4 & 90.5  \\ 
    \hline
    \xpami{TCP} \cite{Cheng_TCP} 
    & 53.7 & 84.3 & 91.0 & 96.3 \\
    \hline
    XQDA \cite{XQDA}
    & 63.2 & 83.9 & 90.0 & 94.9 \\ 
    \hline
    MLAPG \cite{LiaoPSD}
    & 64.2 & 85.5 & 90.8 & 94.9 \\ 
    \hline
    DNS \cite{zhang2016learning}
    & 65.0 & 85.0 & 89.9 & 94.4 \\
    \hline
    \xpami{DGD} \cite{Dropout_reid} 
    & 66.6 & - & - & - \\
    \hline \hline
    \bf{CRAFT-MFA}
    & \bf{74.5} & \bf{91.2} & \bf{94.8} & \bf{97.1} \\
        \hline
  \end{tabular}
  \vspace{-.55cm}
\end{table}

\vspace{.1cm}
\noindent {\em Comparisons on CUHK01 - }
We further conducted the comparative evaluation on
the CUHK01 dataset with the results shown in Table \ref{tab:cuhk01_system}.
It is evident that our CRAFT-MFA method generated the highest person re-id accuracies among
all the compared methods.
\xpami{Specifically,
the best alternative (DGD)
was outperformed notably by our method
with a margin of $7.9\%(=74.5\%-66.6\%)$ rank-1 rate.}
%
Relative to VIPeR,
CRAFT-MFA obtained more
performance increase on CUHK01,
e.g., improving rank-1 rate by $2.5\%(=50.3\%-47.8\%)$ on VIPeR {\em versus}
$7.9\%$ on CUHK01.
This is as expected, {because} person images from VIPeR
are much more challenging for re-id due to
poorer imaging quality, more complex illumination patterns, and
more severe background clutter
(see Figure \ref{examples:figs}(a-b)).
This also validates the generality and capability of the proposed method
in coping with various degrees of person re-id challenges \wsf{when
learning view-specific and view-generic discriminative \zhu{re-id} information.}

\vspace{0.1cm}
\noindent {\bf (II) Person re-id across more than two cameras. }
%
%
Real-world person re-id applications often involve
a surveillance network with many cameras.
It is therefore critical to evaluate the performance of associating people
across a whole camera network, although \wsf{the joint learning and quantification is} largely under-studied in the current literature.
%
In this multi-camera setting, we exploit the generalized CRAFT model
(Eqn. \eqref{eq:general_transformation_extend}) to
learn an adaptive sub-model for each camera view in a principled fashion.
This evaluation was performed on three multi-camera re-id datasets:
{CUHK03 \cite{Deepreid} (with 6 cameras in a university campus),}
QMUL GRID \cite{loy2009multi} (with 8 cameras in an underground station),
and Market-1501 \cite{market} (with 6 cameras near a university supermarket).

\begin{table}[t]
  \centering
  \scriptsize
  \renewcommand{\arraystretch}{0.9}
  \setlength{\tabcolsep}{.3cm}
  \xpami{
  \caption{\footnotesize
    Comparing state-of-the-art methods 
      on {\em CUHK03} \cite{Deepreid}.
  }
  \label{tab:cuhk03_system}
  \vskip -0.4cm
  \centering
  \begin{tabular}{c|c|c|c|c||c}
    \hline
    Rank (\%) & 1 & 5 & 10 & 20 & mAP (\%)\\
    \hline \hline
  SDALF \cite{farenzena2010symmetric} & 4.9 & 21.0 & 31.7 & - & - \\
  \hline
  ITML \cite{ITML} & 5.1 & 17.7 & 2.8.3 & - & - \\
  \hline
  LMNN \cite{LMNN} & 6.3 & 18.7 & 29.0 & - & - \\
  \hline
  eSDC \cite{eSDCocsvm} & 7.7 & 22.0 & 33.0 & - & - \\
  \hline
  KISSME \cite{KISSME} & 11.7 & 33.3 & 48.0 & - & - \\
  \hline
    FPNN \cite{Deepreid} & 19.9 & 50.0 & 64.0 & 78.5 & - \\
    \hline
    BoW \cite{market} & 23.0 & 42.4 & 52.4 & 64.2 & 22.7 \\
    \hline
    CIND-Net \cite{ahmed2015improved} & 45.0 & 76.0 & 83.5 & 93.2 & - \\
    \hline
    XQDA \cite{XQDA} & 46.3 & 78.9 & 88.6 & 94.3 & - \\
    \hline
    LSSCDL \cite{LSSCDL} & 51.2 & 80.8 & 89.6 & - & - \\
    \hline
    MLAPG \cite{LiaoPSD} & 51.2 & 83.6 & 92.1 & 96.9 & - \\
    \hline
    SI-CI \cite{Wang_joint} & 52.2 & 84.9 & 92.4 & 96.7 & - \\
    \hline
    DNS \cite{zhang2016learning} 
    & 53.7 & 83.1 & 93.0 & 94.8  & - \\
    \hline
    S-LSTM \cite{varior2016siamese} & 57.3 & 80.1 & 88.3 & - & 46.3 \\
    \hline
    Gated-SCNN \cite{Gated_SCNN} & 68.1 & 88.1 & 94.6 & - & 58.8 \\
    \hline
    DGD \cite{Dropout_reid} & 75.3 & - & - & - & - \\
    \hline \hline
    {\bf CRAFT-MFA} 
    & {\bf 84.3} & {\bf 97.1} & {\bf 98.3} & {\bf 99.1} & {\bf 72.41} \\
    \hline  
  \end{tabular}
	}
  \vspace{-.55cm}
\end{table}

\vspace{0.1cm}
\xpami{\noindent {\em Comparisons on CUHK03 - }
  We evaluated our approach by comparing the state-of-the-arts on CUHK03 \cite{Deepreid}.
  This evaluation was conducted using detected images.
  It is shown in Table \ref{tab:cuhk03_system} that our method significantly outperformed all competitors, e.g., the top-2 Gated-SCNN/DGD by $16.2\%$/$9.0\%$ at rank-1, respectively.}
  %

\begin{table}[t]
  \scriptsize
  \renewcommand{\arraystretch}{0.9}
  \setlength{\tabcolsep}{.45cm}
  \caption{\footnotesize
    Comparing state-of-the-art methods 
     on {\em QMUL GRID}\cite{loy2009multi}.
  }
  \label{tab:grid_system}
  \vskip -0.4cm
  \centering
  \begin{tabular}{c|c|c|c|c}
    \hline
    Rank (\%) & 1 & 5 & 10 & 20 \\
    \hline \hline
    PRDC \cite{PRDC}
    & 9.7 & 22.0 & 33.0 & 44.3 \\
    \hline
    LCRML \cite{chen2014relevance}
    & 10.7 & 25.8 & 35.0 & 46.5 \\
    \hline
    MRank-PRDC \cite{loy2013person}
    & 11.1 & 26.1 & 35.8 & 46.6\\
    \hline
    MRank-RSVM \cite{loy2013person}
    & 12.2 & 27.8 & 36.3 & 46.6 \\
    \hline
    MtMCML \cite{ma2014person}
    & 14.1 & 34.6 & 45.8 & 59.8 \\
    \hline
    PolyMap \cite{chen2015similarity}
    & 16.3 & 35.8 & 46.0 & 57.6 \\
    \hline
    MLAPG \cite{LiaoPSD}
    & 16.6 & 33.1 & 41.2 & 53.0 \\ 
    \hline
    KEPLER \cite{KEPLER}
    & 18.4 & 39.1 & 50.2 & 61.4 \\
    \hline
    XQDA \cite{XQDA}
    & 19.0 & 42.2 & 52.6 & 62.2 \\ 
    \hline
    \xpami{LSSCDL} \cite{LSSCDL} & {\bf 22.4} & - & 51.3 & 61.2 \\
    \hline
%
    \hline  
    \bf{CRAFT-MFA}
    & \bf{22.4} & \bf{49.9} & \bf{61.8} & \bf{71.7} \\
    \hline  
  \end{tabular}
  \vspace{-.55cm}
\end{table}

\vspace{0.1cm}
\noindent {\em Comparisons on QMUL GRID - }
The re-id results of different methods
on QMUL GRID are presented in Table \ref{tab:grid_system}.
It is found that the proposed CRAFT-MFA method produced
the most accurate results among all competitors, similar to
the observation on CUHK03 above.
In particular, our CRAFT-MFA method outperformed \wsf{clearly} the $2^\text{nd}$ best model \xpami{LSSCDL,
e.g., with similar top-1 matching rate but boosting rank-10 matching from $51.3\%$ to $61.8\%$.}
This justifies the superiority of our CRAFT model and person appearance
feature in a more challenging realistic scenario.

\begin{table}[t]
  \scriptsize
  \renewcommand{\arraystretch}{0.9}
  \setlength{\tabcolsep}{.25cm}
  \caption{\footnotesize
    Comparing state-of-the-art methods
    on {\em Market-1501} \cite{market}.
    ($^+$): the results reported in \cite{zhang2016learning} were utilized.
  }
  \label{tab:Market_system}
  \centering
  \vskip -0.4cm
  \begin{tabular}{c|c|c||c|c}
    \hline
    Query/person & \multicolumn{2}{c||}{Single Query} & \multicolumn{2}{c}{Multiple Query} \\ \hline 
    Metric & rank-1 (\%) & mAP (\%) & rank-1 (\%) & mAP (\%) \\
    \hline \hline
    BoW \cite{market}
    & 34.4 & 14.1
    & 42.6 & 19.5 \\
    \hline
    KISSME$^+$ \cite{KISSME}
    & 40.5 & 19.0
    & - & - \\
    \hline
    MFA$^+$ \cite{MFA}
    & 45.7 & 18.2
    & - & - \\
    \hline
    kLFDA$^+$ \cite{xiao2014KernelMatching}
    & 51.4 & 24.4
    & 52.7 & 27.4 \\
    \hline
    XQDA$^+$ \cite{XQDA}
    & 43.8 & 22.2
    & 54.1 & 28.4 \\
    \hline
    DNS \cite{zhang2016learning}
    & 55.4 & 29.9
    & 68.0 & 41.9 \\
    \hline
    \xpami{S-LSTM} \cite{varior2016siamese} 
    & - & -
    & 61.6 & 35.3 \\
    \hline
    \xpami{Gated-SCNN} \cite{Gated_SCNN}
    & 65.9 & 39.6
    & 76.0 & 48.5 \\
    \hline
        \hline
    {\bf CRAFT-MFA}
    & {\bf 68.7} & {\bf 42.3}
    & {\bf 77.0} & {\bf 50.3} \\
    \hline
  \end{tabular}
  \vspace{-.55cm}
\end{table}

\vspace{0.1cm}
\noindent {\em Comparisons on Market-1501 - }
We compared the performance on Market-1501 %
with these methods:
Bag-of-Words (BoW) based baselines \cite{market},
a Discriminative Null Space (DNS) learning based model \cite{zhang2016learning},
and four metric learning methods
KISSME \cite{KISSME},
MFA \cite{MFA},
kLFDA \cite{xiao2014KernelMatching},
XQDA \cite{XQDA},
\xpami{S-LSTM \cite{varior2016siamese},
Gated-SCNN \cite{Gated_SCNN}.}
We evaluated both the single-query and multi-query
(using multiple probe/query images per person during the deployment stage) settings.
It is evident from Table \ref{tab:Market_system} that
our CRAFT-MFA method outperformed all competitors 
under both single-query and multi-query settings.
By addressing the small sample size problem,
the DNS model achieves more discriminative models than other metric learning algorithms.
However, all these methods focus on learning view-generic discriminative information
whilst overlooking largely useful view-specific knowledge.
Our CRAFT-MFA method effectively overcome this limitation by
encoding camera correlation into an extended feature space
for jointly learning both view-generic and view-specific discriminative information.
\xpami{Additionally, our method benefits from more view change tolerant 
appearance patterns
deeply learned from general auxiliary data source for obtaining
more effective person description,
and surpassed recent deep methods Gated-SCNN and S-LSTM.}
%
All these evidences \wsf{validate} consistently the effectiveness and capability
of the proposed person visual features and cross-view re-id model learning approach
in multiple application scenarios.

%
%

\vspace{0.1cm}
\noindent {\bf (III) Person re-id with multiple feature representations. }
Typically, person re-id can benefit from using multiple different types
of appearance features owing to their complementary effect
(see Table \ref{tab:eval_feat}).
Here, we compared our CRAFT-MFA(+LOMO) method with 
 \zhu{competitive} re-id models using
multiple features.
This comparison is given in Table \ref{tab:fuse}.
It is found that our CRAFT-MFA(+LOMO) method \wsf{notably} outperformed all \wsf{compared} methods
utilizing two or more types of appearance features,
particularly on CUHK01, CUHK03, and Market-1501.
Along with the extensive comparison with single feature based methods above,
these observations further validate the superiority and effectiveness of our proposed method
under varying feature representation cases.
\begin{table}[t]
       \scriptsize
  \renewcommand{\arraystretch}{0.9}
  \setlength{\tabcolsep}{.21cm}
  \caption{\footnotesize
    Comparing state-of-the-art methods using multiple types of appearance feature representations.
    }
  \label{tab:fuse}
  \centering
  \vskip -0.4cm
  \begin{tabular}{c|c|c|c|c}
    \hline
  Dataset & \multicolumn{4}{c}{VIPeR \cite{viper}}
  \\ \hline
    Rank (\%) & 1 & 5 & 10 & 20 \\
    \hline
    Late Fusion \cite{zheng2015query} 
    & 30.2 & 51.6 & 62.4 & 73.8 \\
    \hline
    MLF+LADF \cite{medFilter}
    & 43.4 & 73.0 & 84.9 & 93.7 \\
    \hline
    Metric Ensemble \cite{ensembleReID}
    & 45.9 & 77.5 & 88.9 & 95.8  \\
    \hline
    CVDCA (fusion) \cite{chen2015CVDCA}
    & 47.8 & 76.3 & 86.3 & 94.0 \\
    \hline
    FFN-Net (fusion) \cite{FNNReID2016}
    & 51.1 & 81.0 & 91.4 & {\bf 96.9} \\
    \hline
    DNS (fusion) \cite{zhang2016learning}
    & 51.2 & 82.1 & 90.5 & 95.9 \\
    \hline
    \xpami{SCSP} \cite{SCSP} 
    & 53.5 & {\bf 82.6} & {\bf 91.5} & 96.7\\
    \hline
    \xpami{GOG (fusion)} \cite{GOG}
    & 49.7 & - & 88.7 & 94.5 \\
    \hline
    {\bf CRAFT-MFA}
    & {50.3} & {80.0} & {89.6} & {95.5} \\
        {\bf CRAFT-MFA(+LOMO)}
    & {\bf 54.2} & 82.4 & {\bf 91.5} & {\bf 96.9} \\
        \hline \hline

    Dataset & \multicolumn{4}{c}{CUHK01 \cite{transferREID}} \\
    \hline
    Rank (\%) & 1 & 5 & 10 & 20 \\
    \hline 
    Metric Ensemble \cite{ensembleReID}
    & 53.4 & 76.4 & 84.4 & 90.5  \\ 
    \hline
    FFN-Net (fusion) \cite{FNNReID2016}
    & 55.5 & 78.4 & 83.7 & 92.6 \\
    \hline
    \xpami{GOG (fusion)} \cite{GOG}
    & 67.3 & 86.9 & 91.8 & 95.9 \\
    \hline  
    DNS (fusion) \cite{zhang2016learning}
    & 69.1 & 86.9 & 91.8 & 95.4 \\
    \hline
    \bf{CRAFT-MFA}
    & {74.5} & {91.2} & {94.8} & {97.1} \\
    {\bf CRAFT-MFA(+LOMO)}
         & {\bf 78.8} & {\bf 92.6} & {\bf 95.3} & {\bf 97.8} \\
    \hline \hline

    Dataset & \multicolumn{4}{c}{\xpami{CUHK03} \cite{Deepreid}} \\
    \hline
    Rank (\%) & 1 & 5 & 10 & 20   \\
    \hline 
    DNS (fusion) \cite{zhang2016learning}
    & 54.7 & 84.8 & 94.8 & 95.2  \\
    \hline
    \xpami{GOG} (fusion) \cite{GOG} & 65.5 & 88.4 & 93.7 & - \\
    \hline
    \bf{CRAFT-MFA}
    & 84.3 & 97.1 & 98.3 & 99.1 \\
    {\bf CRAFT-MFA(+LOMO)}
    & {\bf 87.5} & {\bf 97.4} & {\bf 98.7} & {\bf 99.5}  \\
    \hline \hline
    Dataset & \multicolumn{4}{c}{QMUL GRID \cite{loy2009multi}} \\
    \hline
    Rank (\%) & 1 & 5 & 10 & 20 \\
    \hline 
    \xpami{SCSP} \cite{SCSP}
    & 24.2 & 44.6 & 54.1 & 65.2 \\
    \hline
    \xpami{GOG (fusion)} \cite{GOG}
    & 24.7 & 47.0 & 58.4 & 69.0 \\
    \hline
    \bf{CRAFT-MFA}
    & {22.4} & {49.9} & {61.8} & {71.7} \\
    {\bf CRAFT-MFA(+LOMO)}
    & {\bf 26.0} & {\bf 50.6} & {\bf 62.5} & {\bf 73.3} \\
    \hline \hline

    Dataset & \multicolumn{4}{c}{Market-1501 \cite{market}} \\
    \hline
    Query/person & \multicolumn{2}{c|}{Single Query} & \multicolumn{2}{c}{Multiple Query} \\ \hline 
    Metric & \wsf{rank-1} (\%) & mAP (\%) & \wsf{rank-1} (\%) & mAP (\%) \\
    \hline
    BoW(+HS) \cite{market}
    & - & -
    & 47.3 & 21.9 \\
    \hline
    DNS (fusion) \cite{zhang2016learning}
    & 61.0 & 35.7
    & 71.6 & 46.0 \\
    \hline
    \xpami{SCSP} \cite{SCSP} & 51.9 & 26.4 & - & - \\
    \hline
    {\bf CRAFT-MFA}
    & 68.7 & 42.3
    & 77.0 & 50.3 \\
    {\bf CRAFT-MFA(+LOMO)}
    & {\bf 71.8} & {\bf 45.5} & {\bf 79.7} & {\bf 54.3} \\
    \hline
  \end{tabular}
  \vspace{-.55cm}
\end{table}

\xpami{
\subsection{Discussion}
Here, we discuss the performance of HIPHOP 
in case that deep model fine-tuning on
the available labelled target data is performed in prior to feature extraction.
	Our experimental results suggest that this network adaptation 
	can only bring marginal re-id accuracy gain {on our HIPHOP feature,} 
	{e.g., $<1\%$ rank-1 increase for all datasets except Market-1501 ($1.4\%$), although the improvement on using fc6 and fc7 feature maps 
		(which are much inferior to HIPHOP either fine-tuned or not,
		see the supplementary file for more details) 
		is clearer.}
	This validates empirically our argument that lower layer conv filters can be largely task/domain generic and expressive, thus confirming the similar earlier finding \cite{RCNN} particularly in person re-id context. 
	This outcome is reasonable considering that 
	{the amount of training person images could be still insufficient (e.g., 632 on VIPeR, 1940 on CUHK01, 12197 on CUHK03, 250 on QMUL GRID, 12936 on Market-1501) for producing clear benefit}, 
	especially for lower conv layers which tend to be less target task specific \cite{RCNN}.
	Critically, model fine-tuning not only introduces the cost of extra network building complexity but also {\em unfavourably} renders our feature extraction method domain-specific --
	relying on a sufficiently large set of labelled training data in the target domain.
	Consequently, we remain our {\em domain-generic} (i.e.,
	independent of target labelled training data and deployable universally) re-id feature extraction method 
	as our main person image representation generation way.
}


\vspace{-0.5cm}
\section{Conclusion}
\wsff{We have presented a new framework called CRAFT for person re-identification. CRAFT is formed based on the camera correlation aware feature augmentation. It is
capable of jointly learning both view-generic and view-specific discriminative information for
person re-id in a principled manner.}
%
%
Specifically, by creating automatically a camera correlation aware feature space,
view-generic learning algorithms are allowed to induce view-specific sub-models
which simultaneously take into account the shared view-generic discriminative information
so that more reliable re-id models can be produced.
The correlation between per-camera sub-models can be further constrained
by our camera view discrepancy regularization.
%
Beyond the common person re-id between two cameras,
we further extend our CRAFT framework to cope with re-id \wsf{jointly} across a whole
network of more than two cameras.
%
In addition, we develop a general feature extraction method
allowing to construct person appearance representations with
desired view-invariance property by uniquely exploiting less relevant auxiliary object images other than the target re-id training data.
That is, our feature extraction method is universally scalable
and deployable regardless of the accessibility of labelled target training data.
Extensive comparisons against a wide range of state-of-the-art methods
have validated the superiority and advantages of our proposed approach
under both camera pair and camera network based re-id scenarios
on five challenging person re-id benchmarks.


\vspace{-.5cm}
\section*{Acknowledgement}
This work was supported partially by the National Key Research
and Development Program of China (2016YFB1001002),NSFC
(61522115, 61573387, 61661130157), and the RS-Newton Advanced
Fellowship (NA150459), and Guangdong Natural Science
Funds for Distinguished Young Scholar under Grant
S2013050014265. This work was finished when Mr. Chen was
a master student at Sun Yat-sen University. The corresponding
author and principal investigator for this paper is Wei-Shi Zheng.

\ifCLASSOPTIONcaptionsoff
  \newpage
\fi

\bibliographystyle{IEEEtran}
\bibliography{TPAMI_mirror_abbr}

\vspace{-1.2cm}
\begin{IEEEbiography} [{\includegraphics[width=1in,height=1.25in,clip,keepaspectratio] {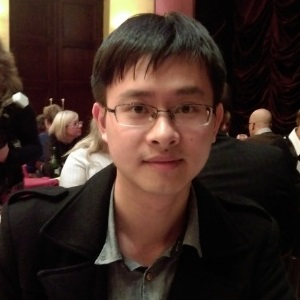}}]{Ying-Cong Chen} received his BEng and Master degree from Sun Yat-sen University in 2013 and 2016 respectively. Now he is a PhD student in the Chinese University of Hong Kong. His research interest includes computer vision and machine learning.
\end{IEEEbiography}

\vspace{-1.2cm}
\begin{IEEEbiography} [{\includegraphics[width=1in,height=1.25in,clip,keepaspectratio]
    {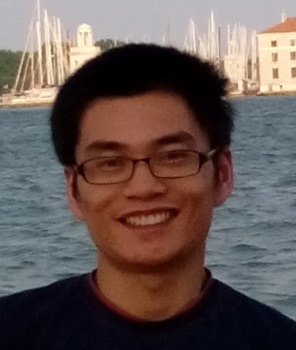}}]{Xiatian Zhu} received his B.Eng. and M.Eng.
  from University of Electronic Science and Technology of China, and
  his Ph.D. (2015) from Queen Mary University of London. 
  He won The Sullivan Doctoral Thesis Prize (2016),
  an annual award representing the best doctoral thesis 
  submitted to a UK University in the field of computer or natural vision.
  His research interests include computer vision,
  pattern recognition and machine learning.
\end{IEEEbiography}

\vspace{-1.2cm}
\begin{IEEEbiography} [{\includegraphics[width=1in,height=1.25in,clip,keepaspectratio]
    {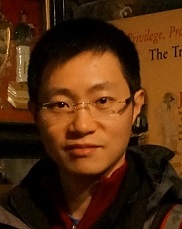}}] {Wei-Shi Zheng} is now a Professor at Sun Yat-sen University. He has now published more than 90 papers, including more than 60 publications in main journals (TPAMI,TIP,PR) and top conferences (ICCV, CVPR,IJCAI). His research interests include person/object association and activity understanding in visual surveillance. He has joined Microsoft Research Asia Young Faculty Visiting Programme. He is a recipient of Excellent Young Scientists Fund of the NSFC, and a recipient of Royal Society-Newton Advanced Fellowship. Homepage: http://isee.sysu.edu.cn/\%7ezhwshi/
\end{IEEEbiography}

\vspace{-1.2cm}
\begin{IEEEbiography} [{\includegraphics[width=1in,height=1.25in,clip,keepaspectratio]
    {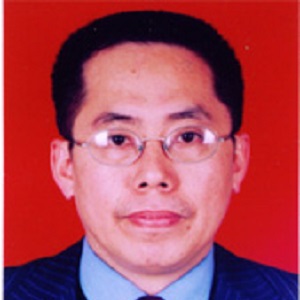}}]{Jian-Huang Lai} is Professor of School of Data and Computer
Science in Sun Yat-sen university. His current research interests are in the areas of digital image processing, pattern recognition, multimedia communication, wavelet and its applications. He has published over 100 scientific papers in the international journals and conferences on image processing and pattern recognition e.g., IEEE TPAMI, IEEE TNN, IEEE TIP, IEEE TSMC (Part B), Pattern Recognition, ICCV, CVPR and ICDM.
\end{IEEEbiography}
%
%
%






\end{document}